\documentclass[runningheads]{llncs}

\usepackage{eccv}

\usepackage{eccvabbrv}

\usepackage{graphicx}
\usepackage{booktabs}
\usepackage{marvosym}

\usepackage[accsupp]{axessibility}  

\usepackage{hyperref}

\usepackage{orcidlink}

\usepackage{mmstyle}
\usepackage{multirow}
\usepackage{makecell}
\usepackage{subcaption}
\usepackage{amsmath}

\usepackage{amssymb}
\usepackage{pifont}

\newcommand{\cmark}{\ding{51}}\xspace%
\newcommand{\xmark}{\ding{55}\xspace}%

\begin{document}


\title{UniTalker: Scaling up Audio-Driven 3D Facial Animation through A Unified Model} 

\titlerunning{UniTalker}

\author{Xiangyu Fan\inst{}\orcidlink{0000-0002-3446-524X} \and
Jiaqi Li\inst{}\orcidlink{0000-0002-0058-0266} \and
Zhiqian Lin\inst{}\orcidlink{0009-0005-5971-1928} \and
Weiye Xiao\inst{}\orcidlink{0000-0003-3015-3609} \and
Lei Yang\inst{}\orcidlink{0000-0002-0571-5924}\textsuperscript{\Letter}}
\authorrunning{X.~Fan et al.}

\institute{SenseTime Research, China \\
\email{\{fanxiangyu, lijiaqi2, linzhiqian, xiaoweiye1, yanglei\}@sensetime.com} 
}

\maketitle

\begin{abstract}
Audio-driven 3D facial animation aims to map input audio to realistic facial motion.
Despite significant progress, limitations arise from inconsistent 3D annotations, restricting previous models to training on specific annotations and thereby constraining the training scale.
In this work, we present \textbf{UniTalker}, a unified model featuring a multi-head architecture designed to effectively leverage datasets with varied annotations.
To enhance training stability and ensure consistency among multi-head outputs, we employ three training strategies, namely, PCA, model warm-up, and pivot identity embedding.
To expand the training scale and diversity, we assemble \textbf{A2F-Bench}, comprising five publicly available datasets and three newly curated datasets.
These datasets contain a wide range of audio domains, covering multilingual speech voices and songs, thereby scaling the training data from commonly employed datasets, typically less than 1 hour, to 18.5 hours.
With a single trained UniTalker model, we achieve substantial lip vertex error reductions of 9.2\% for BIWI dataset and 13.7\% for Vocaset.
Additionally, the pre-trained UniTalker exhibits promise as the foundation model for audio-driven facial animation tasks.
Fine-tuning the pre-trained UniTalker on seen datasets further enhances performance on each dataset, with an average error reduction of 6.3\% on A2F-Bench. Moreover, fine-tuning UniTalker on an unseen dataset with only half the data surpasses prior state-of-the-art models trained on the full dataset.
%
The code and dataset are available at the project page\footnote{Homepage: \url{https://github.com/X-niper/UniTalker}}.

\end{abstract}
\section{Introduction}
\label{sec:intro}

Realistic facial animation synchronized with voice is crucial in human-related animation~\cite{anyi2023dynamic,qing2023story,cai2023digital,siyao2023duolando} and simulation~\cite{black2023bedlam,yang2023synbody,xrfeitoria}.
Traditional methods involve vision-based facial performance capture or labor-intensive handcrafted work by artists.
%
Recent neural network advancements enable expressive 3D facial animation based on vocal audio, categorized as vertex-based and parameter-based models.
%
%
%
Bao~\etal~\cite{bao2023learning} showcased that a personalized model, \ie, a model tailored to an individual and trained with approximately 3,000 utterances, can yield reasonably good results when using the pre-trained speech model~\cite{baevski2020wav2vec,conneau2020unsupervised}. A larger dataset of 10,000 utterances further improved performance~\cite{bao2023learning}. It implies that non-personalized models would require an even larger dataset to attain optimal performance. However, existing datasets like BIWI~\cite{fanelli20103} or Vocaset~\cite{cudeiro2019capture} typically contain less than 1,000 utterances.
To train a robust and generailizable audio-to-face model, an appealing solution is to scale up to a larger dataset by assembling existing datasets, similar to recent studies~\cite{cai2023smpler,zhang2024large}.
Yet, there are two main challenges: \emph{inconsistent data annotation} and \emph{insufficient data variety}.

\begin{figure}[t]
    \centering
    \includegraphics[width=1.0\linewidth]{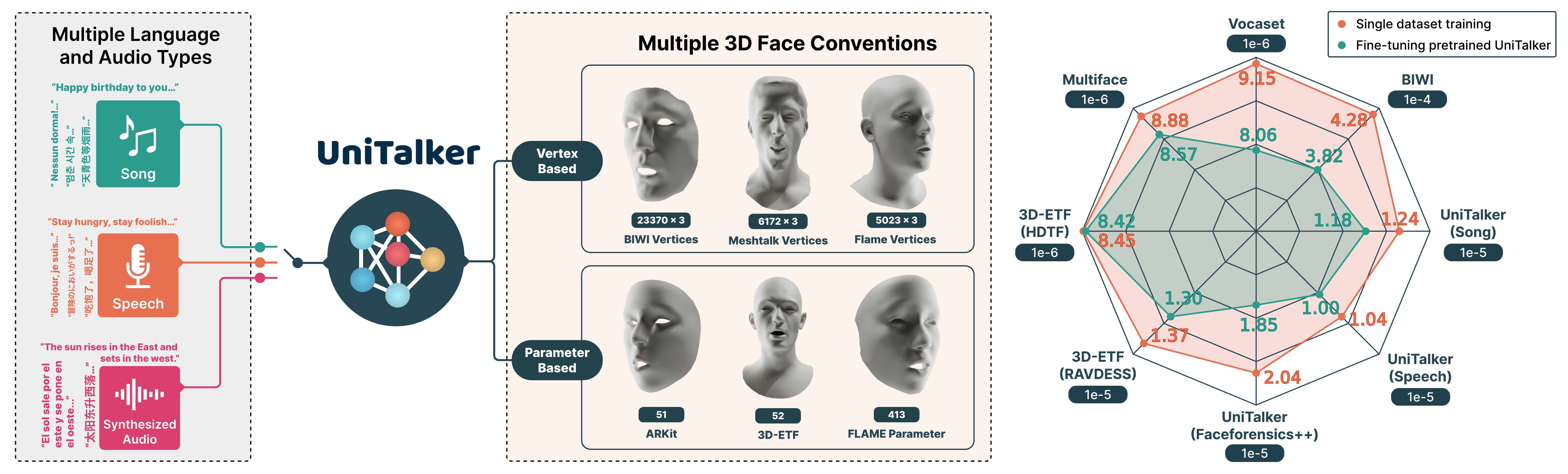}
    \caption{\textbf{Left}: UniTalker aims to learn from diverse datasets in a unified manner.
    It takes multilingual, multi-vocal-type audios as input and outputs various 3D facial annotation conventions simultaneously.
    \textbf{Right}:
    Finetuning UniTalker on each dataset consistently shows lower lip vertex error (LVE) than training the model on the dataset, leading to an average LVE drop of 6.3\%.
    Refer to \cref{tab:LVE_single_all_and_finetune} for comprehensive numerical results.
    }
    \label{fig:teaser}
\end{figure}

To effectively exploit multiple datasets with inconsistent data annotation, we propose UniTalker, a multi-head model that learns from multiple datasets in a unified manner.
However, a straightforward multi-head design faces two primary challenges, notably \emph{training instability} and \emph{dataset bias}.
(1)
As shown in \cref{fig:teaser} and \cref{table:overview_of_datasets}, diverse datasets adhere to distinct annotations.
Vertex-based methods handle thousands of 3D coordinates, while parameter-based methods deal with only a few hundred parameters, leading to different training difficulty.
To address this, we employ Principal Component Analysis for vertex-based annotations to reduce the representation dimension, thus balancing the trainable parameters of different motion decoder heads. 
%
(2) Existing audio-to-face methods often embed speaker identity during training, directly applying it to multiple datasets introduces annotation bias. As there are no shared speakers across datasets with different annotations, dataset bias will leak to the identity embedding module.
Inspired by classifier-free guidance~\cite{ho2022classifier}, we devise Pivot Identity Embedding to mitigate the biases between different motion decoder heads, where a pseudo identity is created and probable to be chosen during training.

%
With the designed unified model, increasing the scale of training necessitates both the quantity and diversity of datasets.
Although there are some publicly available audio-to-face datasets, current datasets predominantly focus on English content and primarily feature a small number of speakers. When dealing with cross-language scenarios, pronunciation and mouth shapes may lack direct counterparts in English (\eg, jiāo in Chinese phonetics).
Furthermore, certain sounds, especially in musical content like American TV shows, require exaggerated mouth movements not commonly found in regular speech.
The lack of such data challenges trained models to accurately reproduce corresponding mouth shapes.
To enrich both sound types and mouth shapes, we curated a multilingual and multi-vocal-type dataset.
The dataset comprises 1.4 hours of Chinese speech and 5.1 hours of multilingual songs.
To increase the diversity of speakers, we annotated the 2D face video dataset FaceForensics++ \cite{rossler1901learning}, contributing additional 3.6 hours of multilingual speech from over 700 individuals.
Combining five existing datasets with three newly curated ones, we assembled \textbf{A2F-Bench}. It contains 934 speakers and 8,654 sequences, with a total duration of 18.53 hours.
\begin{table}[t]\normalsize
\caption{\textbf{Overview of audio-driven 3D facial datasets.} ID refers to dataset identifiers. N denotes the annotation dimension. E, C, M stands for English, Chinese and Multilingual. \#Seq. and \#Subj. means the number of sequences and subjects.}

\centering
\setlength{\tabcolsep}{1mm}
\scalebox{0.6}{
\begin{tabular}{c|c c c c c c c c c c c}
\toprule
Dataset   &  ID & N & GT Type  & Acquisition  & Language  & Audio  & \#Seq.  & Duration & FPS & \#Subj.  & Accessible\\
\hline
BIWI\cite{fanelli20103} & D0                 
& 23,370$\times$3 & Vertices  & 4D Scan                      & E           & Speech & 238   & 0.33h  & 25 & 6   & \cmark\\

Vocaset\cite{cudeiro2019capture} & D1              
& 5,023$\times$3 & Vertices   & 4D Scan                      & E           & Speech & 473   & 0.56h  & 60 &12  & \cmark\\

Multiface(Meshtalk)\cite{wuu2022multiface} & D2  
& 6,172$\times$3 & Vertices  & 4D Scan        & E           & Speech & 612  & 0.67h  & 30 & 13  & \cmark\\

3D-ETF (HDTF)\cite{peng2023emotalk}  & D3  
& 52 & BS  & 3D fitting                 & E           & Speech & 2,039  & 5.49h  & 30 & 141 & \cmark\\

3D-ETF (RAVDESS)\cite{peng2023emotalk}  & D4
& 52 & BS  & 3D fitting        & E           & Speech & 1,440  & 1.48h  & 30 & 24  & \cmark\\

Talkshow\cite{yi2023generating}   & D8          
& 413 & FLAME   & 3D fitting    & E           & Speech & 17,110 & 38.6h & 30 & 4   & \cmark\\

BEAT\cite{liu2022beat}           & D9 
& 52 & BS   & ARKit   & M      & Speech & 2,508   & 76h    & 60 & 30   & \cmark\\

RenderMe-360\cite{pan2023renderme} & -         
& 52 & FLAME & 4D Scan  & C, E  & Speech & 18,000 & 25h     & 30 & 500 & \xmark\\

MMFace4D\cite{wu2023mmface4d}     & -         
& 35,709$\times$3 & Vertices & 4D Scan   & C           & Speech & 35,904    & 36h    & 30 & 431 & \xmark\\

Song2face\cite{iwase2020song2face}  & -          
& 51  & BS & ARKit   & M      & Song   & -         & 1.93h  & - & 7   & \xmark\\

\hline
Ours(Faceforensics++)  & D5  
& 413 & FLAME &  3D fitting & M      & Speech & 1,714     & 3.65h  & 30 & 719 & \cmark\\

Ours(Speech)  & D6              
& 51  & BS & ARKit      & C           & Speech & 789       & 1.24h  & 60 & 8   & \cmark\\

Ours(Song)    & D7              
& 51  & BS & ARKit    & M      & Song   & 1,349     & 5.11h  & 60 & 11  & \cmark\\

\bottomrule
\end{tabular}
}

\label{table:overview_of_datasets}
\end{table}

Leveraging the proposed unified model alongside datasets, a single trained UniTalker achieves lower lip vertex error (LVE) than previous state-of-the-art~\cite{peng2023selftalk}, demonstrating reductions from $4.25$ $\times10^{-4}$ to $3.86$ $\times10^{-4}$ for BIWI and $9.63$ $\times10^{-6}$ $m^2$ to $8.30$ $\times10^{-6}$ $m^2$ for Vocaset.
%
Dataset-specific fine-tuning further enhances the performance and results in an average error reduction of 6.3\% on A2F-Bench. 
%
To demonstrate the generalizability of pre-trained UniTalker, we introduce a practical yet under-explored task, \emph{Annotation Transfer}, which involves transferring to an unseen annotation convention with limited data.
Compared with fine-tuning the commonly adopted audio encoder~\cite{conneau2020unsupervised}, fine-tuning UniTalker requires less than half the data to achieve comparable performance. 

Our contributions are three-folds:
(1) We introduce a multi-head model that integrates diverse datasets and annotation types within a unified framework for 3D facial animation. Our model surpasses existing state-of-the-art with higher accuracy and faster inference speeds.
%
(2) We demonstrate that pre-trained UniTalker can serve as a foundation model for audio-to-face tasks. Fine-tuning on pre-trained UniTalker enhances performance on both seen and unseen annotations, especially when the data scale is limited.
%
(3) We curate A2F-Bench, a large-scale dataset comprising five released high-quality datasets and three newly assembled ones. A2F-Bench enriches the diversity of audio-to-face data and offers a more comprehensive benchmark for audio-to-face methods.

\section{Related Work}
\label{sec:related_work}

%
%
%
%
%
\noindent\textbf{Audio-Driven 3D Facial Animation.}
Early works utilise non-parametric audio features like linear predictive coding (LPC)~\cite{karras2017audio} and Mel Frequency Cepstrum Coefficient (MFCC)~\cite{cudeiro2019capture,wang2011text,shimba2015talking} and regress facial motion from these features with CNN~\cite{karras2017audio}, LSTM~\cite{shimba2015talking} and RNN~\cite{suwajanakorn2017synthesizing}. 
Recent works~\cite{fan2022faceformer,xing2023codetalker,peng2023selftalk,stan2023facediffuser} adopt self-supervised pre-trained speech models like Wav2vec 2.0~\cite{baevski2020wav2vec,conneau2020unsupervised}, Hubert~\cite{hsu2021hubert} and Wavlm~\cite{chen2022wavlm} to extract audio features, greatly enhancing performance and reducing the data requirements. 
Faceformer~\cite{fan2022faceformer} and Codetalker~\cite{xing2023codetalker} model audio-driven facial animation as an auto-regressive problem while Emotalk~\cite{peng2023emotalk} and Selftalk~\cite{peng2023selftalk} model it as regressive.
More recently, diffusion models are incorporated for speech-driven 3D facial animation~\cite{stan2023facediffuser,zhao2024media2face} and improve the diversity of the generated animation. 
Despite achieving realistic facial animation in recent advances, one single model usually focuses on audios of a single domain, \eg, English speech, and outputs one facial animation representation, \eg, vertices of one topology. A unified model is desired that has robust performance in various audio domains, \eg, multilingual speeches and songs, and outputs various 3D representation types, \eg, blendshapes and vertices.
%

\noindent\textbf{Audio-Driven 3D Facial Datasets.}
Existing publicly available audio-visual datasets focus on English speeches and conversations. As listed in \cref{table:overview_of_datasets}, vertex-based datasets that are registered from 4D scans feature short duration and few subjects like BIWI, Vocaset and Multiface. 3D-ETF~\cite{peng2023emotalk} is annotated with pseudo ground truth 52 ARkit blendshape weights from 2D videos~\cite{zhang2021flow,livingstone2018ryerson}. It enlarges the available data scale for the audio-to-face generation task. However, 3D-ETF focuses on English content. The two large-scale datasets, Talkshow and BEAT exhibit audio-annotation misalignment and inaccurate annotation, not suitable for audio-to-face generation. 
RenderMe-360~\cite{pan2023renderme}, MMFace4D~\cite{wu2023mmface4d} and Song2face~\cite{iwase2020song2face} are not publicly accessible.
%
%
In summary, there is a lack of non-English audio-visual data and song-to-face data for academic study. 

\section{Methods}

\subsection{Formulation}
Let $\mM_{1:T}^{i} = (\vm_{1}^{i}, ..., \vm_{T}^{i})$ be a sequence of face motion, where $\vm_{t}^{i}$ denotes the face motion at ${t}$-th frame following the ${i}$-th annotation convention.
For vertex-based annotations, $\vm_{t}^{i} \in \mathbb{R}^{3V}$ denotes the displacement of $V$ vertices at ${t}$-th frame over a neutral-face template.
For parameter-based annotations, $\vm_{t}^{i} \in \mathbb{R}^{P}$ denotes the $P$ parameters at ${t}$-th frame.
Let $\mA_{1:T \cdot d}$ be the input audio, where $d$ is the audio samples aligned with one frame.
%
The goal in this paper can be expressed as follows: Given an input audio $\mA_{1:T \cdot d}$, the model needs to map it into face motion denoted by every desired annotation, \ie, 
$\mM_{1:T}^{i}$, 
$\forall i \leq N$, where $N$ is the number of face annotation types involved in the training process. 

\subsection{Unified Multi-Head Model}
\begin{figure}[ht]
    \centering
    \includegraphics[width=0.88\linewidth]{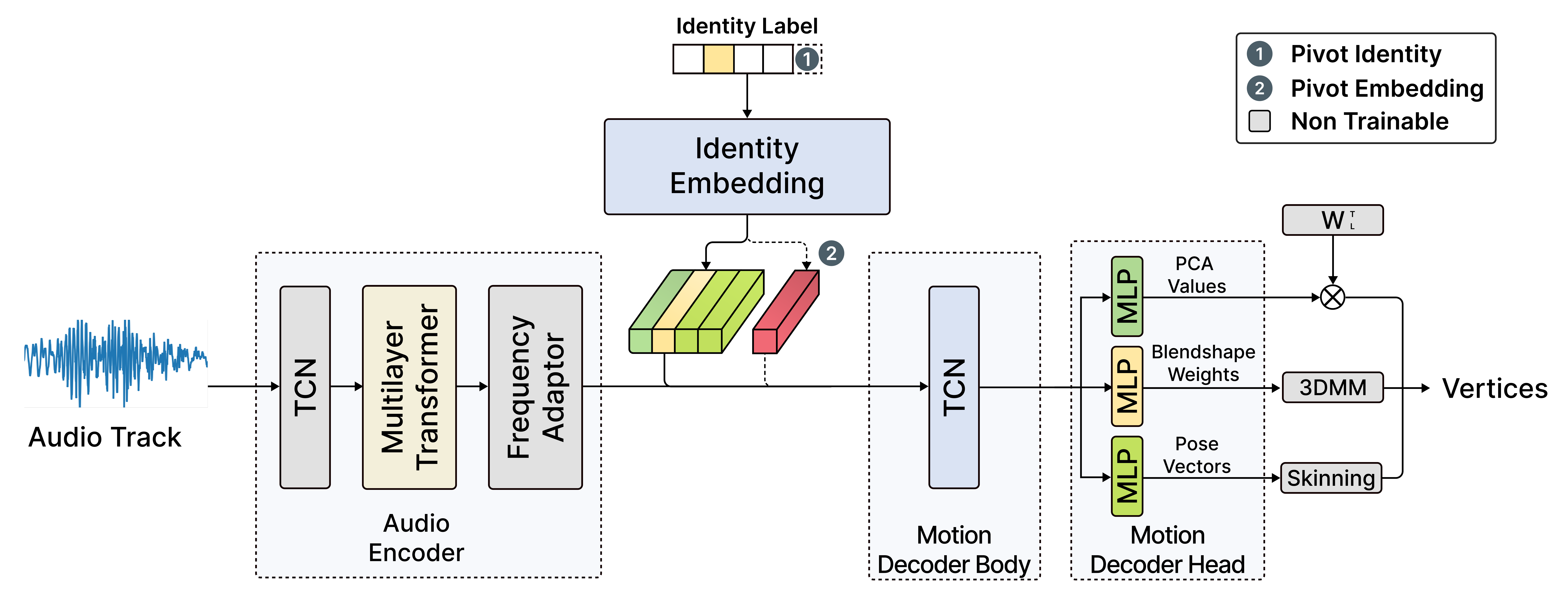}
  \caption{\textbf{UniTalker architecture.} UniTalker adopts vertices PCA to balance the annotation dimension across datasets, uses decoder warm-up to stablize training, and develops a pivot identity embedding to mitigate dataset bias.}
  \label{fig:model_architecture_only_unitalker}
\end{figure}

As shown in \cref{fig:model_architecture_only_unitalker}, our unified multi-head audio-to-face model, namely UniTalker, follows an encoder-decoder architecture.
Given an input audio, the audio encoder initially transforms it into contextualized audio features. Subsequently, the frequency adaptor adapts these audio features via temporal linear interpolation to match the frequency of output face motion. 
%
The motion decoder maps the interpolated audio features into motion hidden states.
Finally, the motion hidden states are decoded onto each annotation through the respective decoder head.

\noindent\textbf{Audio Encoder.}
We adopt the state-of-the-art pre-trained speech model \cite{conneau2020unsupervised,chen2022wavlm} for the audio encoder. Pre-trained audio encoders have been extensively proved to be effective in audio-driven 3D facial animation \cite{fan2022faceformer,xing2023codetalker,bao2023learning,peng2023selftalk,peng2023emotalk,stan2023facediffuser}.
The audio encoder consists of a temporal convolution network (TCN) and a multi-layer transformer encoder. TCN converts the raw audio waveform $\mA_{1:T \cdot d}$ into feature vectors with frequency of 50 Hz and the transformer encodes the feature vectors into contextualized audio representations.

\noindent\textbf{Frequency Adaptor.}
To address varying annotation frequencies across multiple datasets, we incorporate a frequency adaptor into our model. This adaptor performs linear interpolation, aligning audio features from 50 Hz to the frequency of output face motion.
In contrast to prior methods~\cite{fan2022faceformer, xing2023codetalker}, we reposition the frequency adaptor behind the transformer encoder. 
%
%
This adjustment ensures the frequency of the transformer input in training stage is aligned with that in pre-training stage. Hence, the pre-trained weights of the audio encoder are better utilised. 
The result is enhanced convergence and improved model precision, as evidenced in Supplementary Materials.

\noindent\textbf{Non-autoregressive Motion Decoder.}
Faceformer~\cite{fan2022faceformer} and CodeTalker~\cite{xing2023codetalker} have formulated audio-to-face generation as an auto-regression task. 
It involves a motion encoder to project the preceding predicted motion into motion embeddings. 
%
The decoder uses both the motion embeddings and contextualized audio representations to predict the face motion at the next frame.
Other works adopt non-autoregressive models, employing transformer~\cite{bao2023learning, peng2023selftalk} and TCN~\cite{yi2023generating} for the motion decoder.
We observe that removing autoregression from FaceFormer brings 30 times faster inference speed and does not adversely affect precision for either BIWI or Vocaset.
%
UniTalker adopts TCN for the motion decoder as it exhibits better precision for multi-head training. Please refer to Supplementary Materials for detailed results. 

\begin{figure}[tb]
 \begin{minipage}[t]{0.44\textwidth}
    \includegraphics[width=\linewidth]{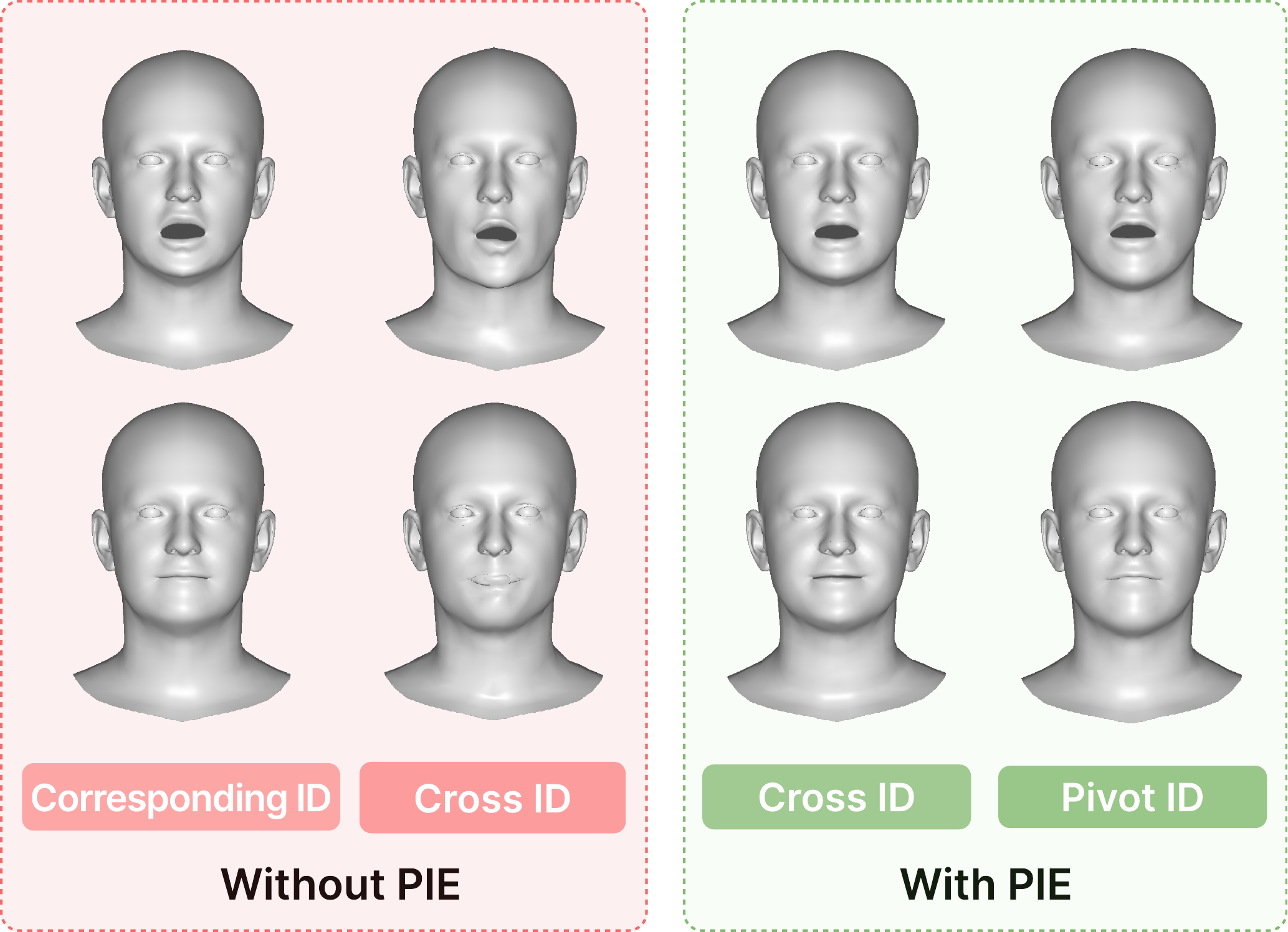}
    \caption{\textbf{Effect of PIE.} Without PIE, the model generates unnatural face motion when input identity and output annotation mismatch.}
    \label{fig:pivot_embedding_compare}
 \end{minipage}
 \hfill\begin{minipage}[t]{0.5\textwidth}
  \includegraphics[width=\linewidth]{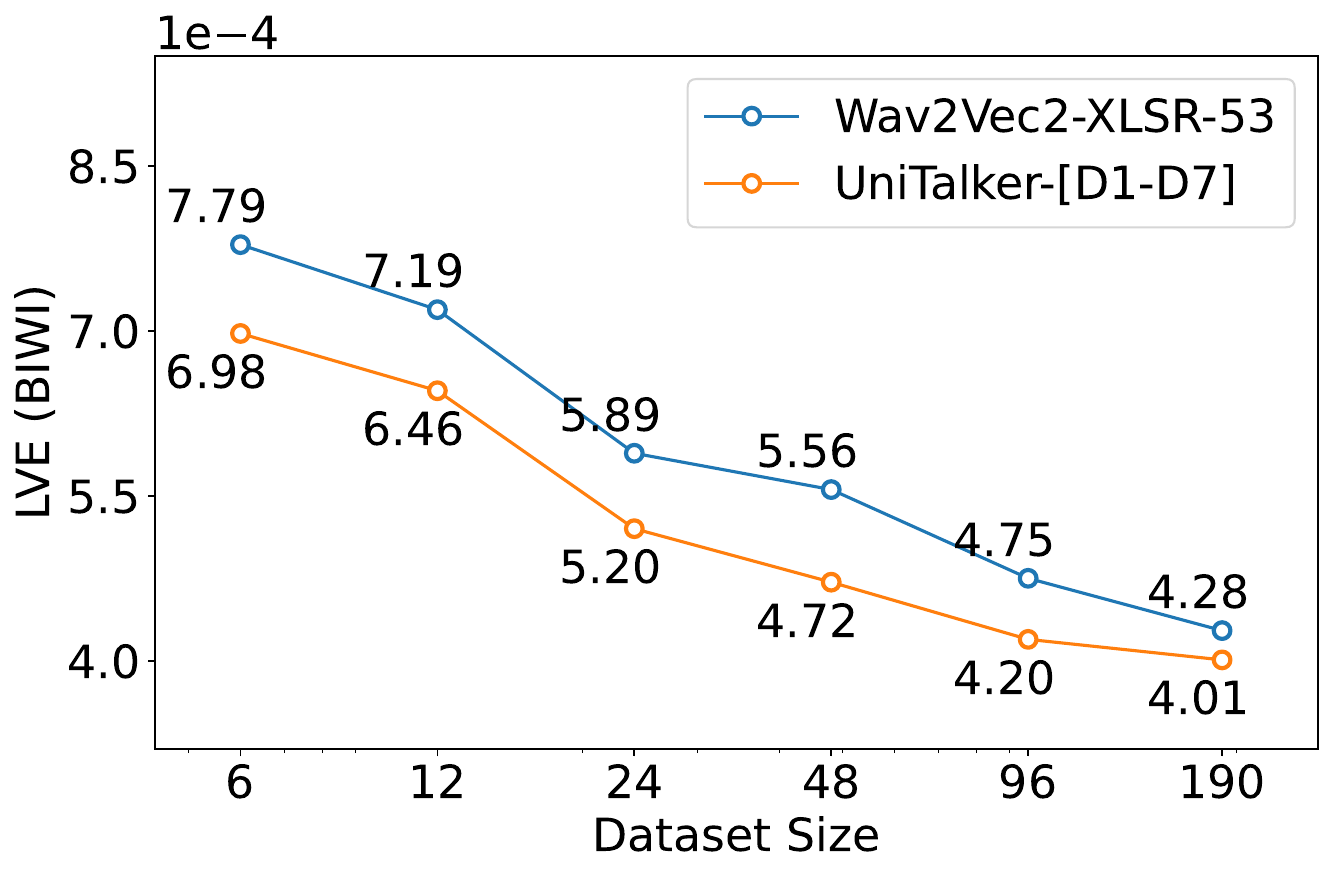}
  \caption{Comparison between finetuning Wav2vec2-xlsr-53~\cite{Wav2Vec2_XLSR_53} and UniTalker-L-[D1-D7] on D0. The x-axis is in \emph{log-scale}.}
  \label{fig:miniBIWI_transfer_compare}
 \end{minipage}
\end{figure}

\noindent\textbf{Identity Embedding.}
To model the speaking styles of different individuals, face motion generation is conditioned on the input identity label, as shown in \cref{fig:model_architecture_only_unitalker}.
The speakers in different datasets are exclusive to each other, implying that each motion decoder head is trained within a specific subset of speakers and audios. 
As a result, the decoder head of one annotation does not necessarily output natural face motion when the input identity label and audio belong to another annotation.
\cref{fig:pivot_embedding_compare} shows that the model generates satisfactory face motion only when conditioned on an identity label from the corresponding annotation. Unnatural face motion, \eg, weird mouth shape and self-intersection may be generated when input identity and motion decoder head mismatch (Cross ID inference). 
Inspired by classifier-free diffusion guidance~\cite{ho2022classifier}, we propose Pivot Identity Embedding (PIE) to mitigate the annotation biases. Specifically, we introduce an additional pivot identity that does not belong to any datasets, as shown in \cref{fig:model_architecture_only_unitalker}. During training, we replace the ground truth (GT) identity label with this pivot identity label with a probability of 10\%. 
%
\cref{fig:pivot_embedding_compare} shows that UniTalker exhibits the ability to generate satisfactory face motion regardless of the identity label used for conditioning.

\subsection{Unified Multi-Head Training}
 
\begin{figure}[tb]
  \centering
  \begin{subfigure}{0.24\linewidth}
    \includegraphics[width=\linewidth]{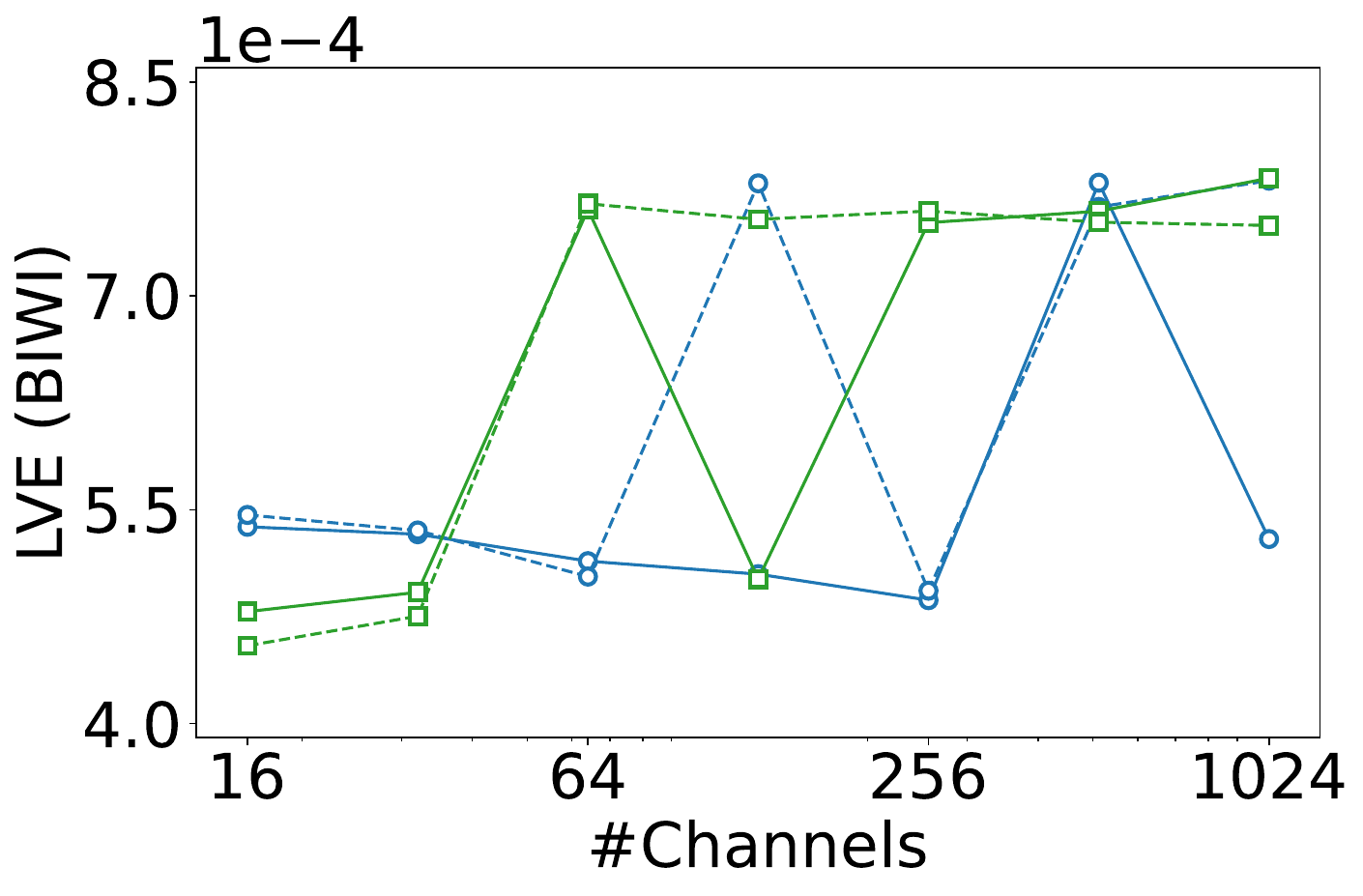}
    \label{subfig:BIWI_wo_PCA_wo_DW}
  \end{subfigure}
  \begin{subfigure}{0.24\linewidth}
    \includegraphics[width=\linewidth]{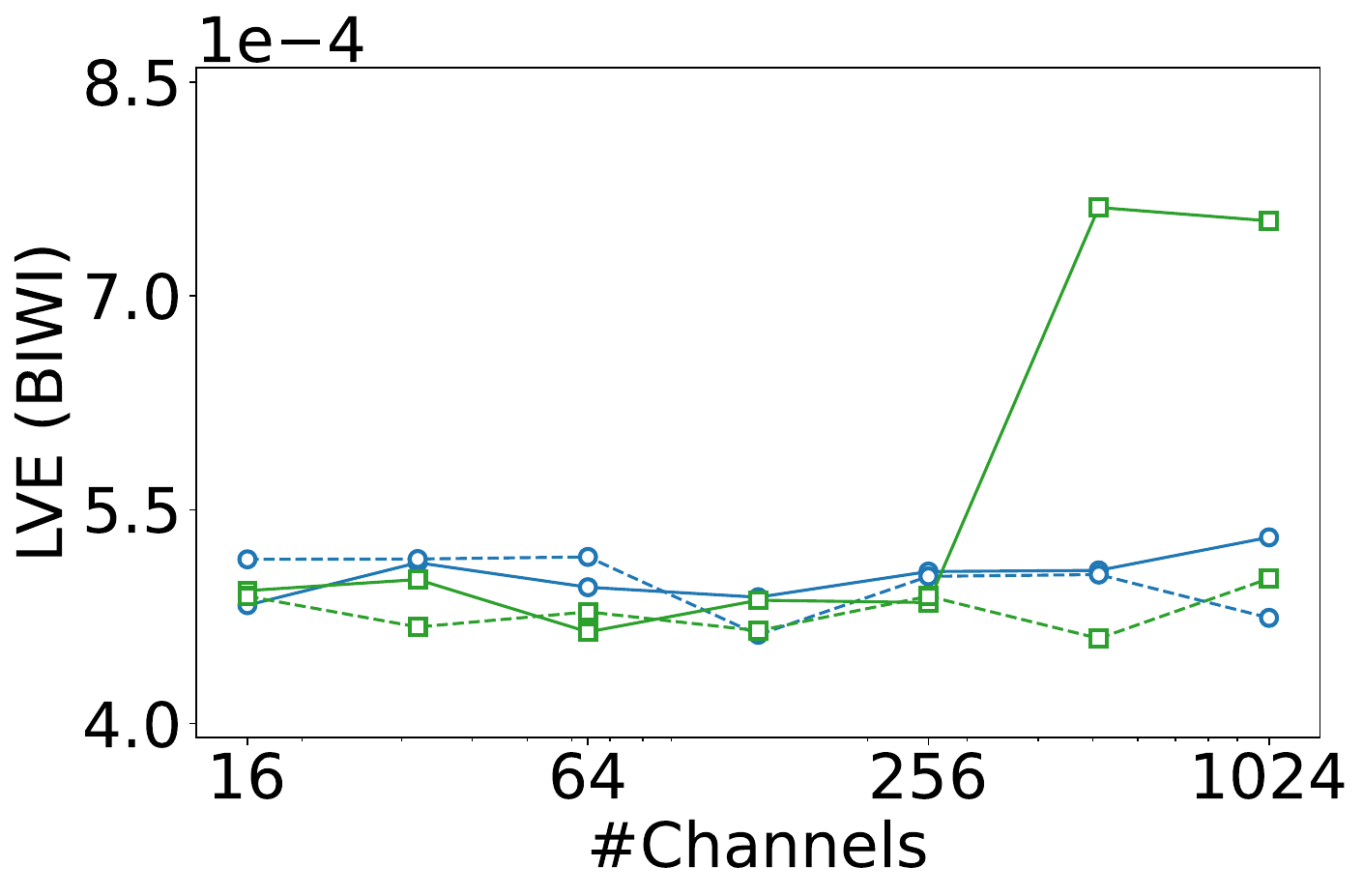}
    \label{subfig:BIWI_wo_DW}
  \end{subfigure}
  \begin{subfigure}{0.24\linewidth}
    \includegraphics[width=\linewidth]{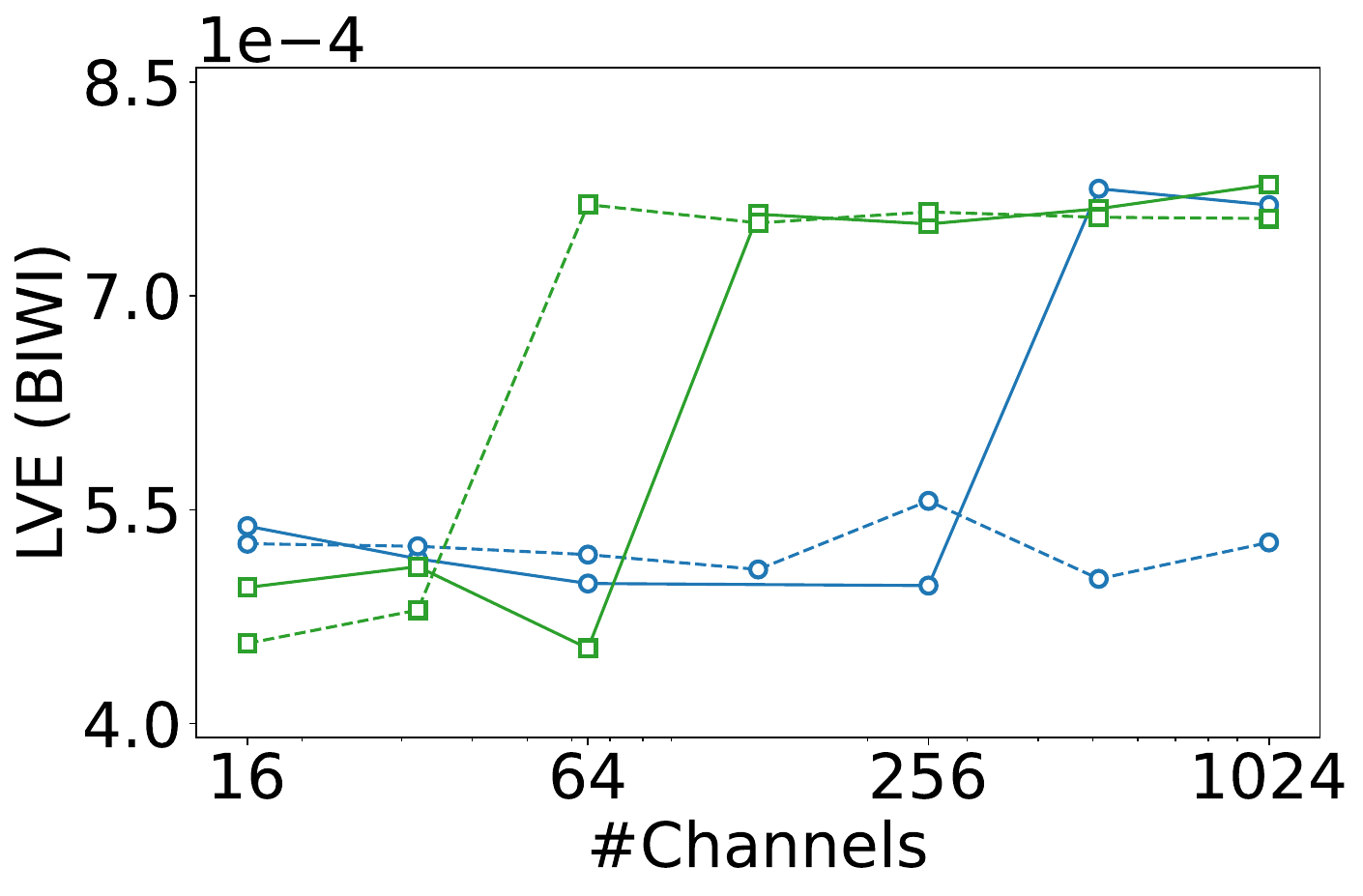}
    \label{subfig:BIWI_wo_PCA}
  \end{subfigure}
  \begin{subfigure}{0.24\linewidth}
    \includegraphics[width=\linewidth]{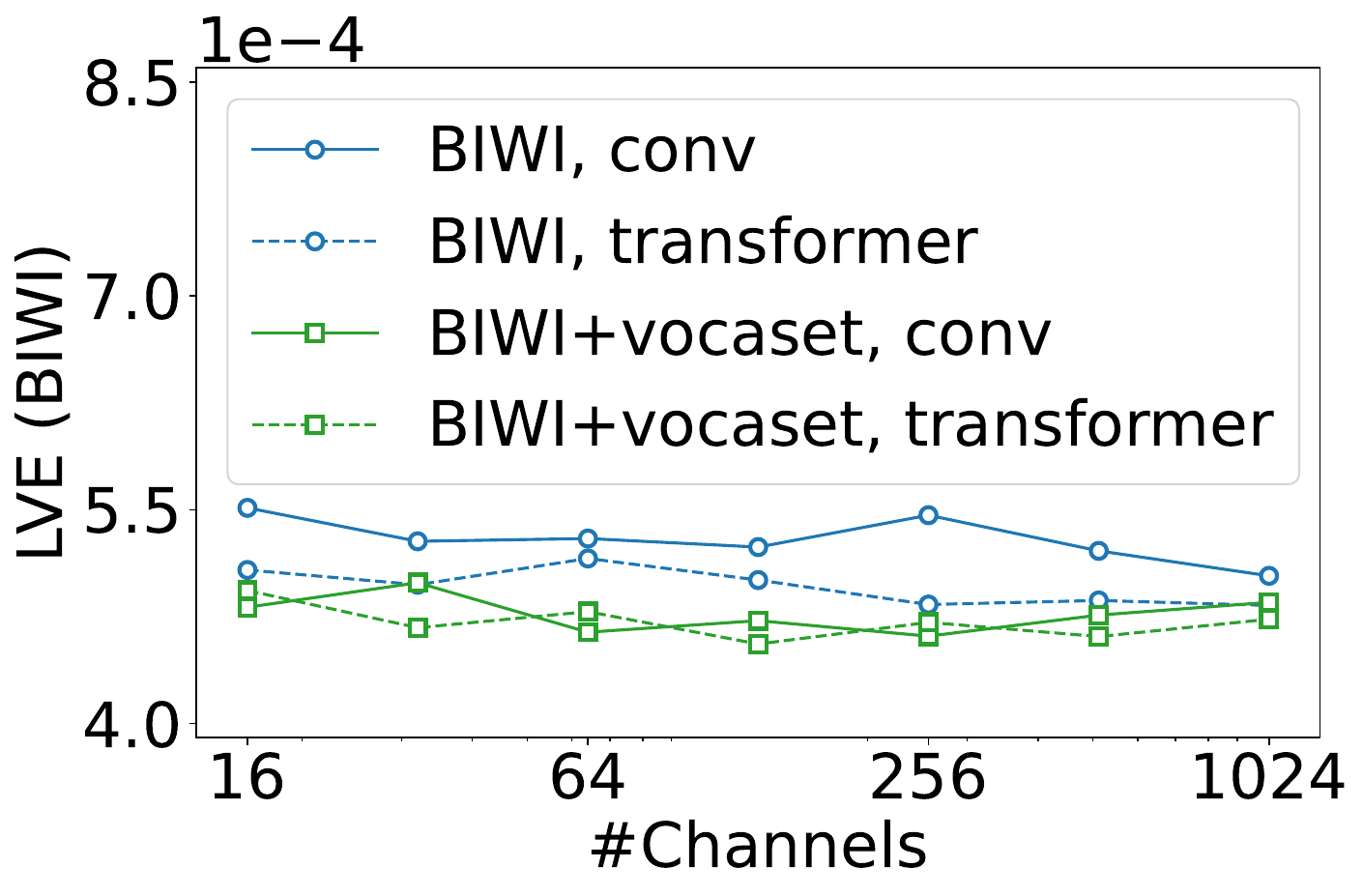}
    \label{subfig:BIWI_with_PCA_and_DW}
  \end{subfigure}
  \begin{subfigure}{0.24\linewidth}
    \includegraphics[width=\linewidth]{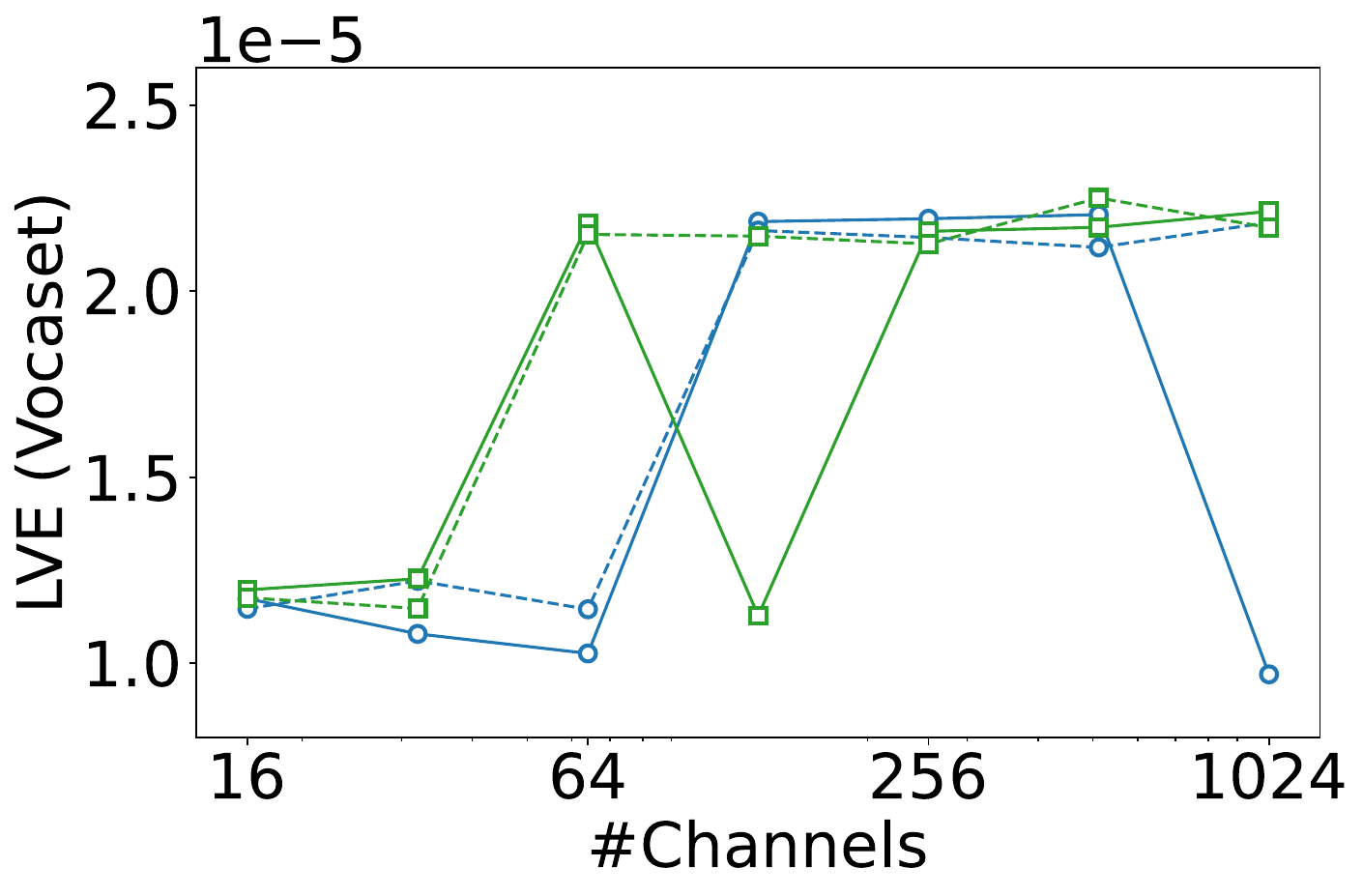}
    \caption{w.o. PCA, w.o. DW}
    \label{subfig:vocaset_wo_PCA_wo_DW}
  \end{subfigure}
  \begin{subfigure}{0.24\linewidth}
    \includegraphics[width=\linewidth]{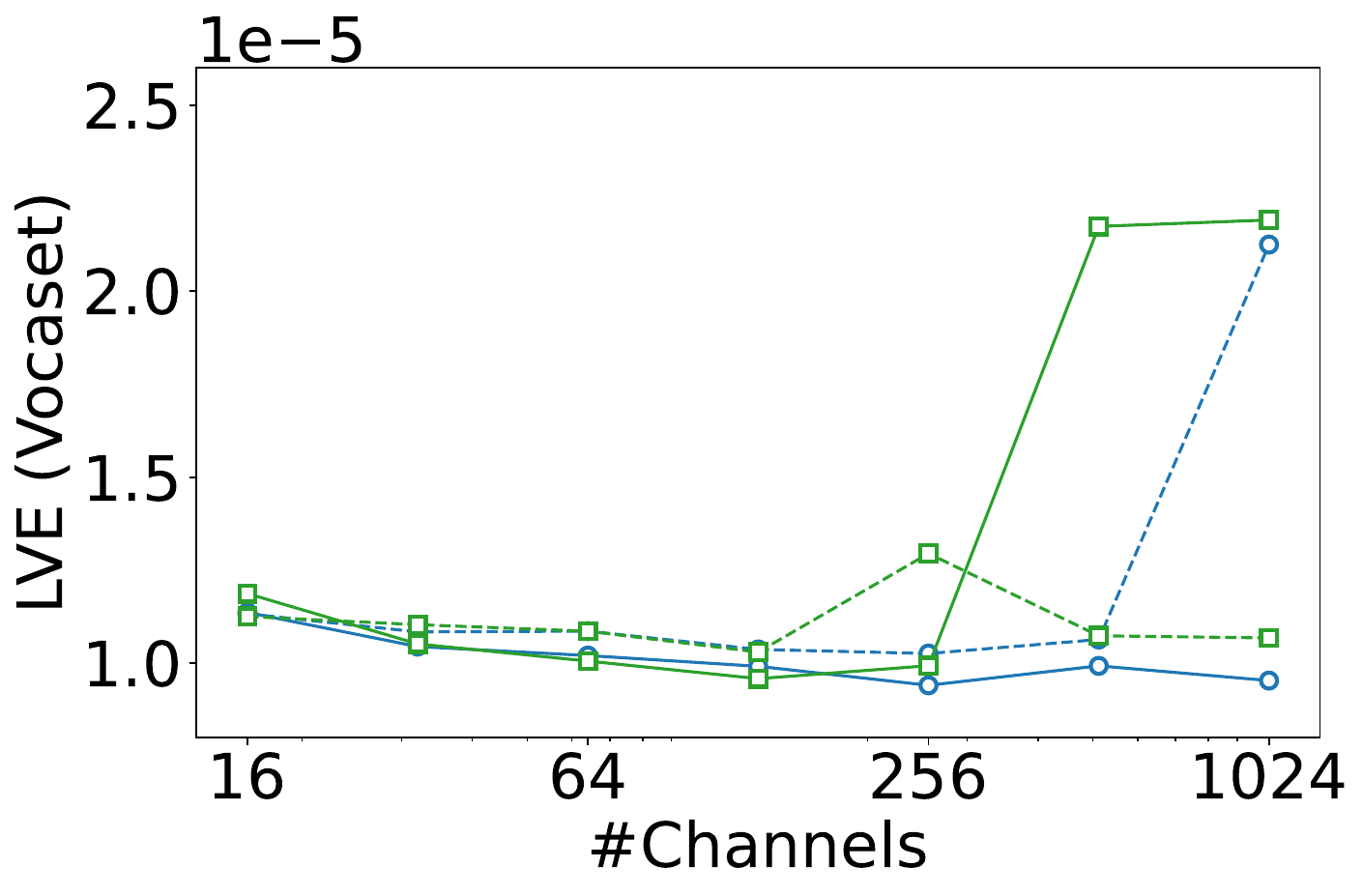}
    \caption{w.o. DW}
    \label{subfig:vocaset_wo_DW}
  \end{subfigure}
  \begin{subfigure}{0.24\linewidth}
    \includegraphics[width=\linewidth]{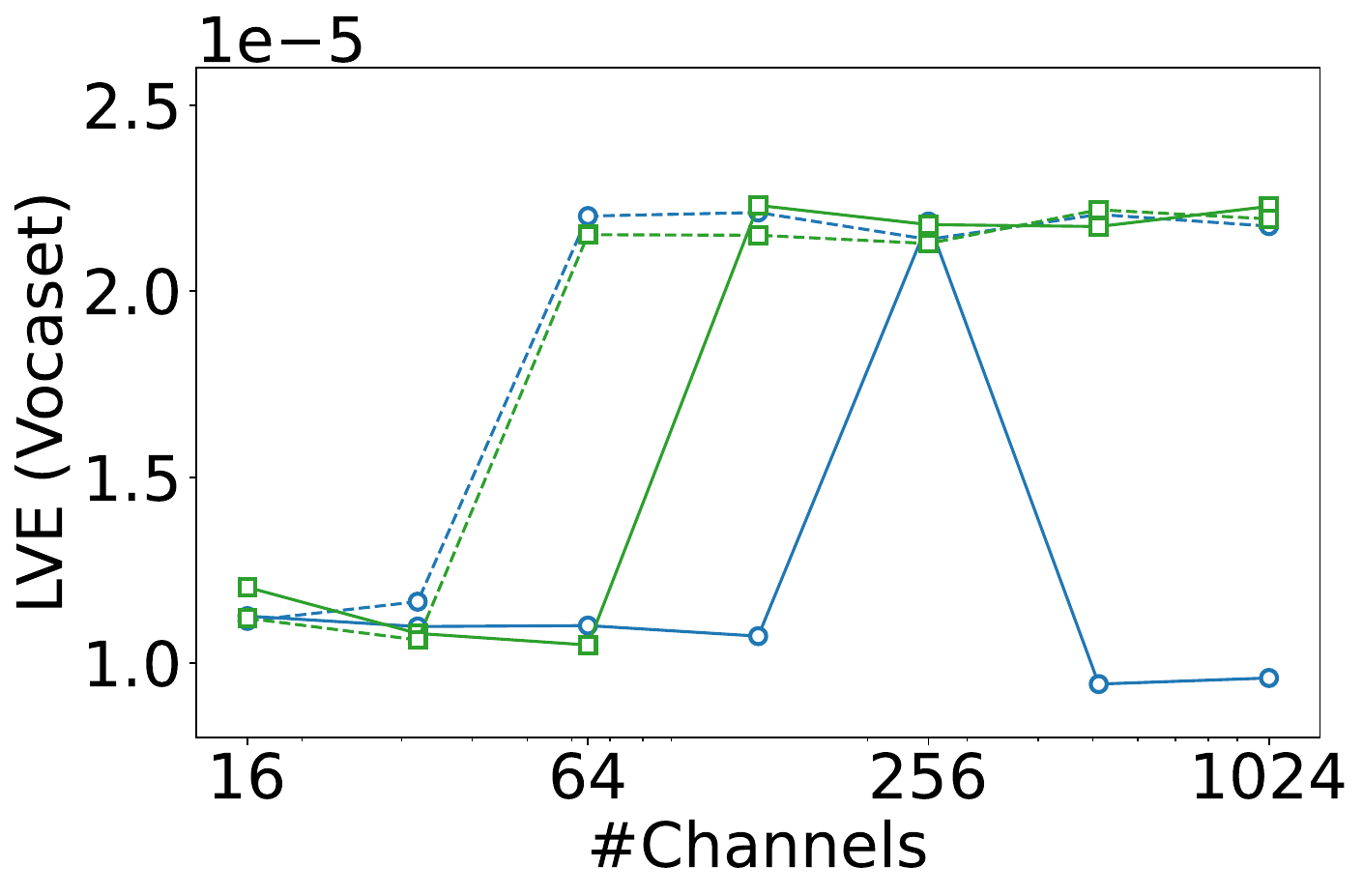}
    \caption{w.o. PCA}
    \label{subfig:vocaset_wo_PCA}
  \end{subfigure}
  \begin{subfigure}{0.24\linewidth}
    \includegraphics[width=\linewidth]{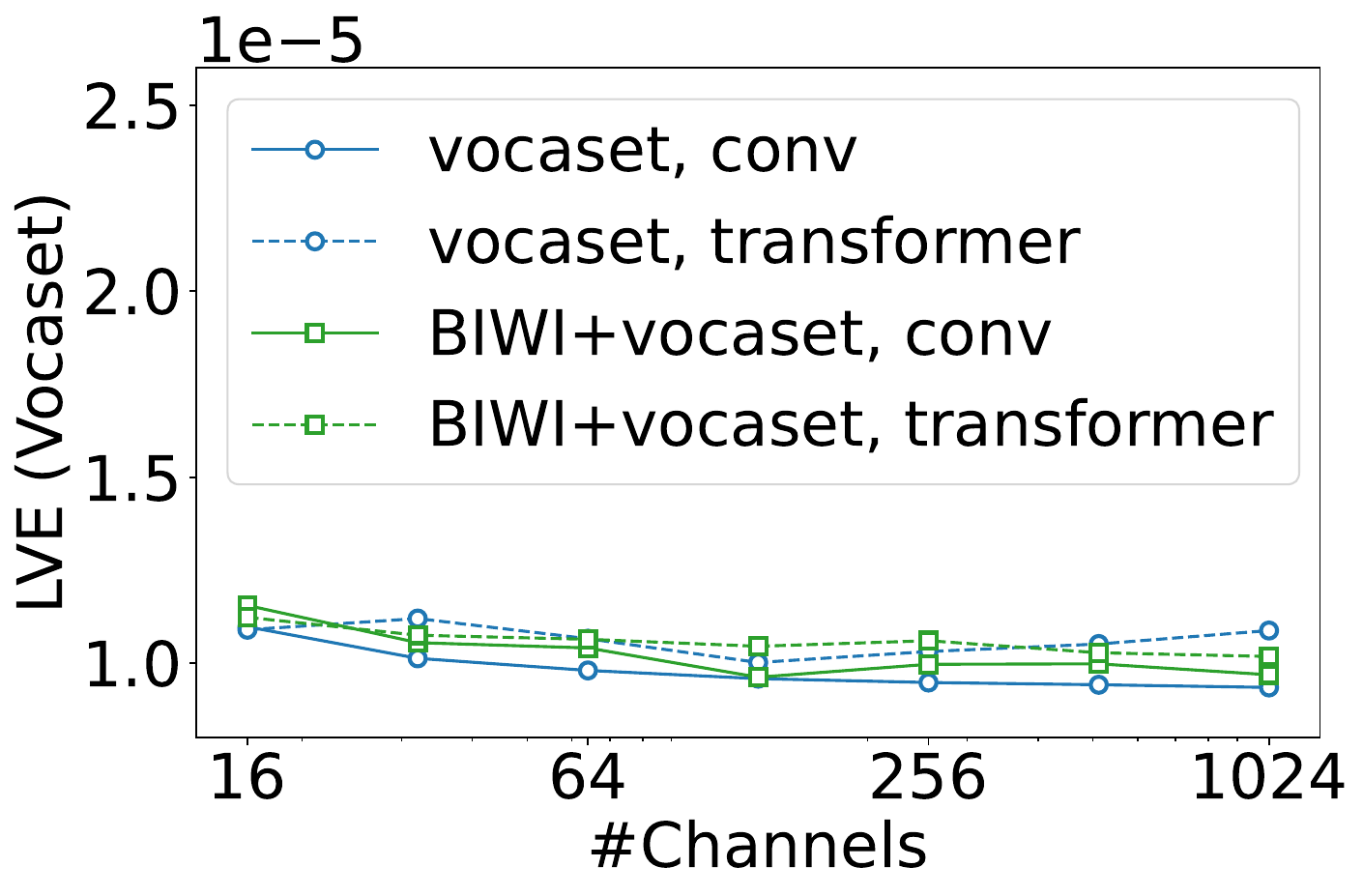}
    \caption{w. PCA, w. DW}
    \label{subfig:vocaset_with_PCA_and_DW}
  \end{subfigure}
  \caption{\textbf{The effect of PCA and DW.}
  LVE values are evaluated on test set at 100th epoch.
  Training with both PCA and DW ensures training stability across various
  settings. Removing either strategy harms training robustness.}
  \label{fig:effect_of_PCA_and_DW}
\end{figure}

\noindent\textbf{Improving Training Stability.}\label{sec:PCA_DW}
A vanilla multi-head model (shown in Supplementary Materials) associates each annotation convention with one output head. However, the vanilla multi-head model fails to gain advantages from increased data size. 
%
We hypothesize that the difference in annotation dimensions results in different difficulties of training convergence.
For example, BIWI and Vocaset possess 23,370 and 5,023 vertices, respectively.
Previous studies~\cite{fan2022faceformer,xing2023codetalker} have chosen distinct hyperparameters for these datasets.
We conducted systematical experiments for the two datasets, across different decoder channels and decoder architectures, using the same audio encoder adopted in FaceFormer~\cite{fan2022faceformer}.
As shown in \cref{subfig:vocaset_wo_PCA_wo_DW}, the model precision is highly related to the hyperparameters and the optimal hyperparameters for the two datasets are different. 

To train the multi-head model stably, we employ \emph{Principal Component Analysis (PCA)} for each vertex-based annotation. This process reduces the output dimension and maintain consistent output head dimensions for each vertex-based annotation. 
%
%
Restricted by memory limit, we employ Incremental Principal Components Analysis (I-PCA)~\cite{ross2008incremental} as an approximation of PCA. It reduces the dimension of motion representation from ${3V}$ to $L=512$, where $V$ denotes the vertex number and $L$ denotes the number of the preserved principle components. Each decoder head for vertices is then replaced with a decoder head for PCA values. The PCA values $\hat{\vy}_{PCA}$ and vertices $\hat{\vy}_{v}$ are linked through the PCA components $\mW_L^T$, according to Eq.~\eqref{eq:pca_values_to_vertices}.
\begin{equation}
    \hat{\vy}_{v} = \hat{\vy}_{PCA} \times \mW_L^T \label{eq:pca_values_to_vertices}
\end{equation}
%

We further stabilize the multi-head training by adopting a two-stage training scheme~\cite{wu2023speech}. In the first stage, we freeze the weights of the pre-trained audio encoder and only update the weights of the decoder. This stage, named \emph{Decoder Warm-up (DW)}, gradually aligns the convergence state of the randomly initialized decoder to that of the pre-trained audio encoder. In the second stage, both the audio encoder and the motion decoder are updated simultaneously.

\cref{fig:effect_of_PCA_and_DW} illustrates the effect of PCA and DW. With both strategies, the model converges across various scenarios, including training on single and multiple datasets, employing either TCN or transformer architectures for motion decoder, and covering a wide range of decoder channel options.
%
\cref{subfig:vocaset_wo_PCA_wo_DW} shows that the vanilla model collapses in many settings and the optimal setting for BIWI and Vocaset is different. Removing either PCA or DW will deteriorate training stability, especially for multi-dataset training, as shown in \cref{subfig:vocaset_wo_DW} and \cref{subfig:vocaset_wo_PCA}.

\noindent\textbf{Training Loss.}
As shown in \cref{fig:model_architecture_only_unitalker}, the model predicts PCA values $\hat{\vy}_{PCA}$ for vertex-based annotations, blendshape weights and pose vectors $\hat{\vy}_{\theta}$ for parameter-based annotations. 
We can derive vertices $\hat{\vy}_{v}$ for every annotation through differentiable computation. 
We apply mean squared error (MSE) on both the model output and the derived vertices, as indicated by \cref{eq:loss_equation},

\begin{equation}\label{eq:loss_equation}
\mathcal{L} = l(\hat{\vy}_{v}, \vy_{v}) + \alpha \cdot l(\hat{\vy}_{PCA}, \vy_{PCA}) + \beta \cdot l(\hat{\vy}_{\theta}, \vy_{\theta}),
\end{equation}
where $\alpha=0.01$ and $\beta=0.0001$ in our training.
%

%
\subsection{UniTalker as a Foundation Model}
Our UniTalker model could output different types of face annotations. In real-wold scenarios, new annotation conventions often arise, and the available data is typically limited. In such cases, the UniTalker model needs to be transferred onto the new annotations. 
Previous works~\cite{fan2022faceformer,stan2023facediffuser,xing2023codetalker} adopts pre-trained audio encoders to decrease the data requirement. In this work, we replace the weights of audio encoder with the weights of pre-trained UniTalker, and find that UniTalker can further decrease half of the data requirement on unseen datasets, as evidenced in Fig.~\ref{fig:miniBIWI_transfer_compare} and discussed in \cref{sec:Results-Annotation-Transfer}.
Additionally, we randomly select only one sequence from Vocaset, which is less than 10 seconds. We fine-tune UniTalker with limited trainable parameters on this single sequence and find that the tuned model can still output satisfactory results (see Supplementary Materials). Note that Vocaset is excluded from the pre-training datasets in this experiment. 

\section{Experiments and Results}\label{sec:Experiments and Results}

\subsection{Datasets: A2F-Bench}
\cref{table:overview_of_datasets} presents a summary of the datasets.
To assemble \textbf{A2F-Bench}, we first select five widely used 3D audio-visual datasets, namely 
BIWI~\cite{fanelli20103}, Vocaset~\cite{cudeiro2019capture}, Multiface~\cite{wuu2022multiface}, 3D-ETF-HDTF~\cite{peng2023emotalk} and 3D-ETF-RAVDESS~\cite{peng2023emotalk}. 
%
%
Additionally, to increase the number of speakers, we clean the multilingual 2D faceforensics++ dataset~\cite{rossler1901learning} and label speaker's faces with FLAME~\cite{li2017learning} parameters using 3D face reconstruction~\cite{dad3dheads,lin2022high}.
To enhance the model's proficiency with non-English speech and songs, we collect a dataset consisting of speeches from eight native Chinese speakers and a dataset comprising multilingual songs from eleven professional singers and label them with ARKit blendshape weights.
We have made experiments on larger datasets like BEAT~\cite{liu2022beat} and TalkShow~\cite{yi2023generating}, and find they exhibit audio-annotation misalignment and inaccurate annotation. 
Hence, they are not included in UniTalker training.
For the sake of simplicity, we refer to each dataset as D0, D1, and so on as in~\cref{table:overview_of_datasets}.
Consistent with previous studies \cite{fan2022faceformer,xing2023codetalker,peng2023emotalk}, we downsample annotations originally collected at 60 fps to 30 fps. BIWI is maintained at 25 fps.
The assembled \textbf{A2F-Bench} consists of 934 speakers and 8,654 sequences, with a total duration of 18.53 hours, featuring diverse sound types and mouth shapes.
%
Refer to Supplementary Materials for detailed dataset description.
\subsection{Implementation Details}
We adopt two multilingual pre-trained audio encoders for UniTalker, \ie, Wavlm-base-plus~\cite{WavLM_Base_Plus} for UniTalker-Base model and Wav2vec2-xlsr-53~\cite{Wav2Vec2_XLSR_53} for UniTalker-Large model. The effect of the audio encoder is detailed in \cref{sec:audio encoder ablation}. UniTalker refers to UniTalker-Large by default, unless explicitly stated. 
We train each version of the model on both individual datasets and A2F-Bench.
For instance, UniTalker-B-[D0] refers to UniTalker-Base trained on BIWI dataset.
UniTalker-B-[D0-D7] and UniTalker-L-[D0-D7] refers to Unitalker-Base and UniTalker-Large trained on the entire A2F-Bench, respectively.
%
We use Adam optimizer with a constant learning rate of 0.0001. We train 100 epochs for each model. It takes 2 days to train UniTalker-L-[D0-D7] on a single NVIDIA V100.

\subsection{Comparison with Prior Works}\label{sec:compare with prior works}

\noindent\textbf{Quantitative Evaluation.}
We compare UniTalker with four methods: FaceFormer~\cite{fan2022faceformer}, CodeTalker~\cite{xing2023codetalker}, SelfTalk~\cite{peng2023selftalk} and FaceDiffuser~\cite{stan2023facediffuser}.
FaceFormer and CodeTalker adopt Wav2vec2-base-960h~\cite{Wav2Vec2_Base_960h} as their audio encoder. Both methods employ autoregressive decoder and exhibit slow inference. SelfTalk adopts Wav2vec2-large-xlsr-53-English \cite{grosman2021xlsr53-large-english} as the audio encoder. FaceDiffuser adopts Hubert-base-ls960~\cite{facebook/hubert-base-ls960} as the audio encoder.
The inference on FaceDiffuser is extremely slow since it adopts the diffusion mechanism and its inference scheduler has 500 steps. 
In case of BIWI, we directly evaluate their released models.
For Vocaset, we retrain and test these methods using their official codebases, as they did not report the quantitative results.

We adopt lip vertex error (LVE) to measure lip synchronization, which is commonly used in prior works~\cite{fan2022faceformer,xing2023codetalker,stan2023facediffuser}. LVE is computed as the average over all frames of maximal L2 error of the lip vertices to the ground truth. 
Following \cite{stan2023facediffuser}, we measure mean vertex error by computing the mean Euclidean distance \wrt the ground truth across all vertices (MVE) and across the upper face (UFVE). 
Following \cite{xing2023codetalker}, we adopt upper-face dynamics deviation (FDD) to measure the variation of upper facial dynamics for a motion sequence in comparison with that of the ground truth.
%
%
We also list the trainable parameters and inference time of a 10-seconds audio on a single NVIDIA V100. 
\begin{table}[tb]
\centering
\caption{Quantitative results on BIWI-Test-A and VOCA-Test.
Best values are bolded.
}
\label{tab:LVE_MVE_FDD_biwi_vocaset_camera_ready}
\scriptsize 
\begin{tabular}{c  c  c c c c c c }
\toprule

\multirow{2}{*}{\textbf{Dataset}} & \multirow{2}{*}{\textbf{Method}} & \textbf{LVE} $\downarrow$ & \textbf{MVE} $\downarrow$ & \textbf{UFVE} $\downarrow$ & \textbf{FDD} $\downarrow$ &\textbf{Params} & \textbf{Time} \\

& &  $\times 10^{-4}$ & $\times 10^{-3}$ & $\times 10^{-3}$ & $\times 10^{-5}$ & $M$ & $s$ \\
\midrule
\multirow{7}{*}{BIWI} &FaceFormer  & 4.9836   & 7.2750  & 6.9081 & 4.0062  &  109     & 0.705 \\
& CodeTalker  & 4.7914  & 7.3784  & 7.0050 & 4.2147 &  561   & 4.4  \\
& SelfTalk  & 4.2485  & 6.9152  & 6.5428 & \textbf{3.5851} &  539    & 0.071  \\
& FaceDiffuser  & 4.2985 & 6.8088  & 6.6220 &  3.9101 &  189  & 16.50 \\
& UniTalker-B-[D0] & 4.3681  & 6.8948  & 6.6277 & 4.6789 & 92 & 0.024  \\
& UniTalker-B-[D0-D7] & 4.0804 & 6.6458  & 6.3774 & 5.0438 & 92 & 0.024  \\
& UniTalker-L-[D0-D7] & \textbf{3.8587} & \textbf{6.4166} & \textbf{6.1483} & 5.2307 & 313 & 0.054  \\
\midrule
\multirow{2}{*}{} & \multirow{2}{*}{} & \textbf{LVE} $\downarrow$ & \textbf{MVE} $\downarrow$ & \textbf{UFVE} $\downarrow$ & \textbf{FDD} $\downarrow$ &\textbf{Params} & \textbf{Time} \\
& &  { $\times 10^{-5}$ $m^{2}$ } & { $\times 10^{-3} m$ } & { $\times 10^{-3} m$ } & { $\times 10^{-7} m^{2}$ } & { $M$ } & { $s$ } \\
\midrule
\multirow{7}{*}{Vocaset} &FaceFormer  & 1.1696 & 0.6364 & 0.4972 & 2.4812  & 92      & 0.624\\
& CodeTalker  & 1.1182 & 0.5750 & 0.4708 & 1.2594   & 315      & 3.464 \\
& SelfTalk  & 0.9626 & 0.5665 & 0.4805 & \textbf{1.0511}  & 450     & 0.053 \\
& FaceDiffuser   & 0.9684 & 0.5768 & 0.4772 & 1.7335 & 89     & 13.08\\
& UniTalker-B-[D1] & 0.9381 & 0.5695 & 0.4829 & 1.2115  & 92 & 0.022 \\
& UniTalker-B-[D0-D7] & \textbf{0.8136} & \textbf{0.5338} & \textbf{0.4494} & 1.3962 & 92 & 0.022 \\
& UniTalker-L-[D0-D7] & 0.8303 & 0.5524 & 0.4756 & 1.5206  & 313 & 0.053\\
\bottomrule
\end{tabular}
\end{table}

\begin{figure}[tb]

\begin{minipage}[]{0.35\textwidth}
  \centering
  \begin{subfigure}{\linewidth}
    \includegraphics[width=\linewidth]{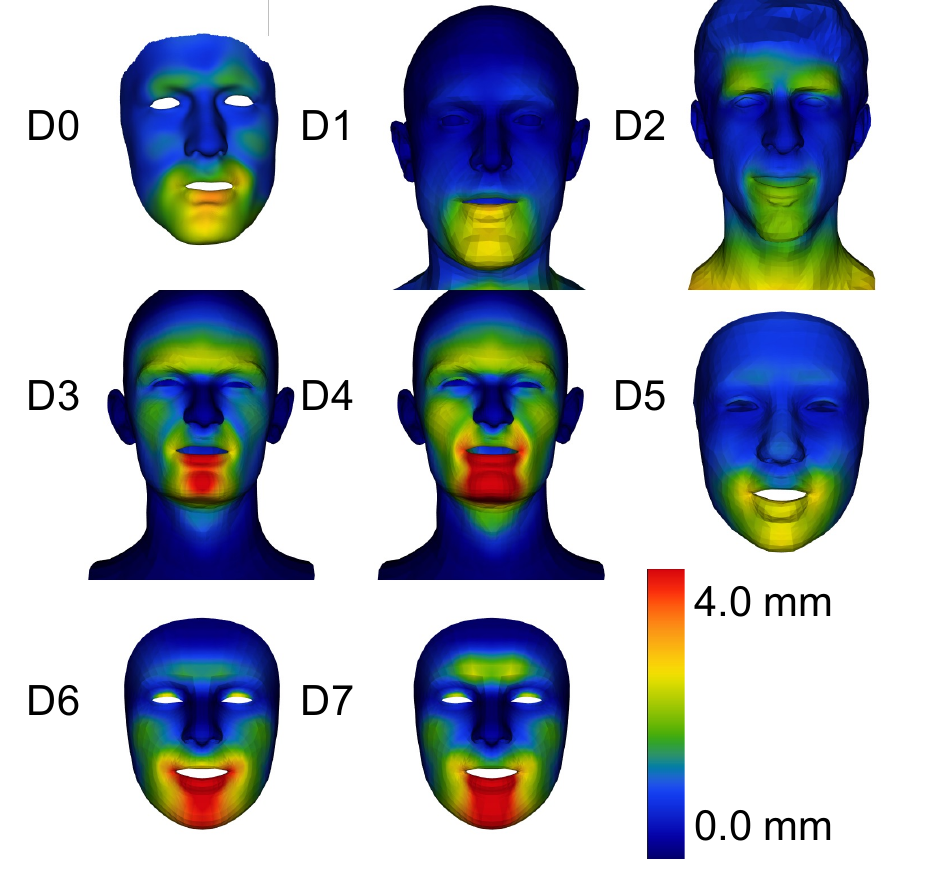}
    \caption{}
    \label{subfig:heat_map_dataset}
  \end{subfigure}
\end{minipage}
\begin{minipage}[]{0.6\textwidth}
  \begin{subfigure}{\linewidth}
    \includegraphics[width=\linewidth]{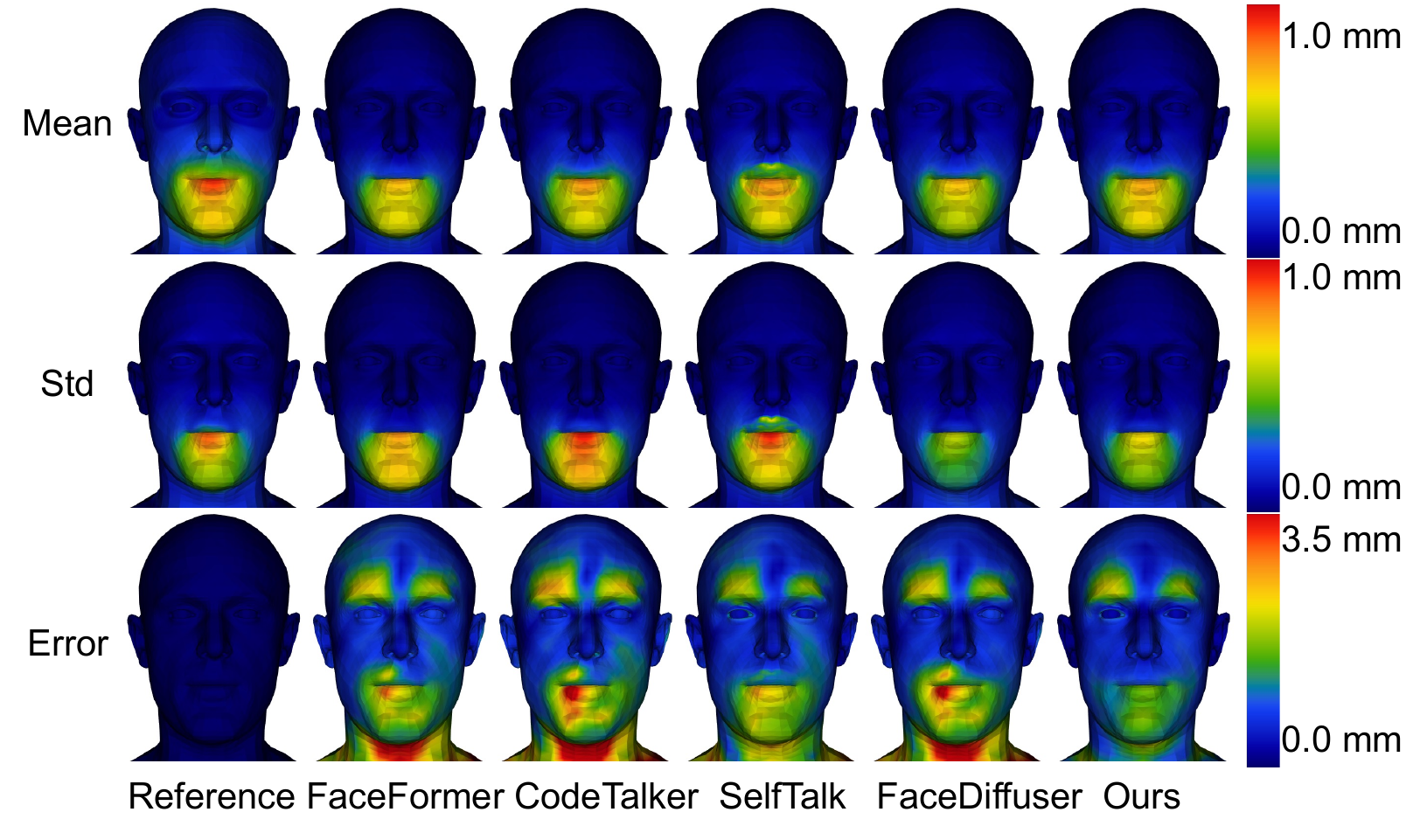}
    \caption{}
    \label{subfig:heat_map_mean_std_error}
  \end{subfigure}
\end{minipage}
\caption{\textbf{(a)} The standard deviation of facial motion within each training set.  The upper face of D1(Vocaset) shows little motion variation and is close to static. \textbf{(b)} The temporal statistics (mean and standard deviation) of adjacent-frame motion variation and the mean of per-frame predicted-to-GT Euclidean distance within a sequence.
}
\label{fig:error_map_ab}
\end{figure}

According to \cref{tab:LVE_MVE_FDD_biwi_vocaset_camera_ready}, UniTalker-B-[D0] and UniTalker-B-[D1] shows lower LVE, than FaceFormer and CodeTalker on BIWI and Vocaset, respectively. With the addition of more training data, UniTalker-B-[D0-D7] get a performance bonus for both datasets and beats all prior works on both datasets in regards to LVE, MVE and UFVE, with less parameters and much faster inference speed.
UniTalker-L-[D0-D7] push LVE, MVE and UFVE even lower on BIWI.
Compared with prior state-of-the-art model, \ie, SelfTalk~\cite{peng2023selftalk}, UniTalker-B-[D0-D7] leads to LVE reductions of 4.0\% for BIWI and 15.5\% for Vocaset. UniTalker-L-[D0-D7] leads to reductions of 9.2\% for BIWI and 13.7\% for Vocaset. 
SelfTalk shows the best FDD on both datasets, indicating the best prediction of statistics of facial motion velocity. Note that although FDD and UFVE are computed over the same upper face region, they show inconsistent results. We argue that UFVE better reflects the temporal consistency with the ground truth. \eg, for $t$$\in$${[0, 2\pi]}$, ${std(cos(t))-std(sin( t))=0}$, implies FDD = 0 and ${\int_0^{2\pi} \lVert cos(t) - sin(t)\rVert_2 dt = 4 \sqrt{2} }$ indicates large UFVE. Notably, diverse data leads to worse FDD, possibly due to the increased diversity of facial motion statistics as shown in \cref{subfig:heat_map_dataset}. For instance, D1 (Vocaset) shows little motion variation in the upper face region while D4 (3DETF-RAVDESS) and D7 (Multilingual Songs) exhibit rich motion variation. At inference, the model trained on diverse datasets tends to predict average motion variation due to the weak correlation between audio and the motion of upper face.
%

\noindent\textbf{Qualitative Evaluation.}
Corroborating the quantitative results above, we plot the mean and standard deviation of the motion velocity, and the mean of the Euclidean distance between the generated sequences and the reference sequence.
According to \cref{subfig:heat_map_mean_std_error}, SelfTalk predicts closest velocity mean and standard deviation maps to the ground truth, which is consistent with the FDD order in \cref{tab:LVE_MVE_FDD_biwi_vocaset_camera_ready}. The error map indicates UniTalker gain the best precision, which is consistent with the LVE, MVE and UFVE results.
Interestingly, prior works show much larger error in the neck part than UniTalker.

\noindent\textbf{User Study.}
We conducted user study to qualitativly compare UniTalker with prior works, FaceFormer, CodeTalker and SelfTalk. FaceDiffuser~\cite{stan2023facediffuser} reported worse qualitative results than FaceFormer and CodeTalker, so it is not selected for comparison. 
Our selected audios for user study cover a wide range of scenarios, including different languages, audio types, emotional expressions, and audio sources (human voices and generated audios from text-to-speech models). 
In our Supplementary Materials, we provide a demo video to illustrate the performance of UniTalker under these scenarios.
For each comparison pair, the output from UniTalker and its competitors were randomly placed at left or right. Users participating in the study were asked to answer three questions for every comparison pair: (1) which side appears more realistic, (2) which side demonstrates better lip synchronization with the audio, and (3) which side more effectively conveys the emotion in the audio.
We collected 868 answers, with 308, 280 and 280 responses compared with  Faceformer, CodeTalker and SelfTalk, respectively.
~\cref{tab:user_study} indicates that UniTalker achieves higher support rate across all three questions. 
\begin{table}[tb]
\centering
\caption{The support rate for UniTalker over its competitors.}
\label{tab:user_study}
\scriptsize 
\begin{tabular}{l | c c c}
\toprule
\textbf{Method} & \textbf{ Realistic } &  \textbf{ Lip Sync } & \textbf{ Emotion } \\
\midrule
{ Ours \vs FaceFormer } &  74.7\% & 76.6\% & 78.2\% \\
{ Ours \vs CodeTalker } & 71.8\% & 77.1\% & 80.7\% \\
{ Ours \vs SelfTalker } & 72.5\% & 75.0\% & 82.1\% \\
\bottomrule
\end{tabular}
\end{table}

\subsection{Comparison With Data Preprocessing}
\label{sec:compare_with_preprocessing}
To train on multiple datasets, one straightforward approach is to preprocess different annotations in the datasets into one unified annotation through either 3D morphable model~\cite{li2017learning} fitting or mesh retopology~\cite{amberg2007optimal}. 
While both methods require pre-selected corresponding facial keypoints, UniTalker does not.
Moreover, the preprocessing approach limits future data expansion.
When a new released dataset adheres to a different annotation, preprocessing approach needs to convert the new annotation into the required format. While for UniTalker, one can simply plug new decoder heads into UniTalker and train it with existing datasets or solely with new ones, avoiding retopology or fitting process.

To quantitatively compare the preprocessing approach with UniTalker, we preprocess all the annotations in [D0-D7] into FLAME vertices, namely [D0-D7]-FLAME,  and train a one-head model on this dataset.
Specifically, for vertex-based datasets like D0 (BIWI) and D2 (Multiface), we convert the vertices into FLAME topology through standard retopology method. The error between the original vertices and converted vertices is evaluated with chamfer distance and has an average value of 0.2 $mm$. 
For D3, D4, D6 and D7, we convert the ARkit blendshape weights into FLAME vertices with the aid of the released blendshape~\cite{liu2023emage} with ARkit semantics and FLAME topology.
For D5, we convert FLAME parameters into vertices using FLAME model~\cite{li2017learning}.
\begin{table}[t]
\caption{
We compare LVE of UniTalker and that of data preprocessing approach, under different training dataset settings. The LVE values are evaluated on D1(VOCA-Test) and expressed in ${10^{-6}}$ ${m^{2}}$. The first row indicates the training datasets.} 
\centering 
\label{table:compare_to_preprocess} 
\scriptsize 
\begin{tabular}{l | c  c  c  c  c  c  c  c  }
\toprule
\textbf{Method} & \textbf{ D1 } & \textbf{ D0-D1 } & \textbf{ D0-D2 }  & \textbf{ D0-D3 } & \textbf{ D0-D4 }  &  \textbf{ D0-D5 }  &  \textbf{ D0-D6 }  &  \textbf{ D0-D7 } \\
\midrule
{ Preprocessing } & 9.1528 &        9.4856 &         8.2400 &   \textbf{8.0779} &     
     8.4730 & 8.7049 & 8.4748 & 8.7532 \\
{ UniTalker }    & 9.1528 & \textbf{8.7353} & \textbf{7.9243} &     8.4495 &   \textbf{8.2336} & \textbf{8.0785} & \textbf{8.4192} & \textbf{8.3035}  \\
\bottomrule 
\end{tabular} 
\end{table}

The one-head model only outputs annotation of FLAME vertices. We compare the performance on D1 (VOCA-Test), which originally has FLAME topology. \cref{table:compare_to_preprocess} shows that UniTalker achieves lower LVE in most dataset settings than the one-head model trained on [D0-D7]-FLAME. Interestingly, the lowest LVE occurs in different dataset settings for these two approaches. \cref{table:compare_to_preprocess} reveals that the unified training framework does take advantages of the multi-head design. UniTalker is not only versatile due to its multi-annotation output, but also shows better precision than data preprocessing approach.
\subsection{Effect of Scaled-up Datasets}
We train UniTalker on each individual dataset and get eight models, denoted as L-[D*]. We evaluate LVE of each model on its corresponding test set. After that, we evaluate LVE of UniTalker-[D0-D7] on every test set. As shown in \cref{tab:LVE_single_all_and_finetune}, the one UniTalker model beats the individual models on most dataset. For small-scale datasets like BIWI and Vocaset, UniTalker leads to over 9\% decrease in LVE.
However, the performance improvement is not achieved on all datasets.
As the audio domains differ largely among A2F-Bench, UniTalker needs to balance the performance across datasets. For D3 (3D-ETF-HDTF), which already contains 5.49 hours of audios, UniTalker does not lead to better precision. For D6 (Chinese speech), UniTalker results in higher LVE because the proportion of Chinese speeches in A2F-Bench is small. 
\begin{table}[tb]
\scriptsize
\centering
\caption{Quantitative comparison between single dataset training and mixed dataset training. The metric is LVE. L-[D*] denotes the eight individual models trained on each dataset. L-[D0-D7] denotes UniTalker-Large trained on A2F-Bench. L-FT denotes the eight models finetuned from L-[D0-D7].
LVE is in ${10^{-4}}$ for D0, ${10^{-6}}$ ${m^{2}}$ for D1-D3 and ${10^{-5}}$ ${m^{2}}$ for D4-D7.
}
\label{tab:LVE_single_all_and_finetune}
\begin{tabular}{l | c c c c c c c c}
\toprule
\multirow{2}{*}{\textbf{Method}} & \textbf{D0}  & \textbf{D1}  & \textbf{D2} & \textbf{D3} & \textbf{D4} & \textbf{D5}  & \textbf{D6} &\textbf{D7} \\
 & 0.33h  & 0.56h & 0.67h & 5.49h & 1.48h & 3.65h & 1.24h & 5.11h \\

\midrule
L-[D*] &  4.279 & 9.153 & 8.881 & 8.445 & 1.370 & 2.040 & 1.043 & 1.235 \\
L-[D0-D7] & 3.859\textcolor{Green}{$\downarrow_{9.8\%}$}   & 8.303\textcolor{Green}{$\downarrow_{9.3\%}$}     & 8.648\textcolor{Green}{$\downarrow_{2.6\%}$} & 8.991\textcolor{Red}{$\uparrow_{6.5\%}$} & 1.326\textcolor{Green}{$\downarrow_{3.2\%}$} & 2.056\textcolor{Red}{$\uparrow_{0.8\%}$} & 1.145\textcolor{Red}{$\uparrow_{9.7\%}$} & 1.211\textcolor{Green}{$\downarrow_{1.9\%}$} \\
L-FT  & 3.816\textcolor{Green}{$\downarrow_{11\%}$}   & 8.060\textcolor{Green}{$\downarrow_{12\%}$}     & 8.56\textcolor{Green}{$\downarrow_{3.5\%}$} & 8.417\textcolor{Green}{$\downarrow_{0.3\%}$} & 1.30\textcolor{Green}{$\downarrow_{5.2\%}$} & 1.848\textcolor{Green}{$\downarrow_{9.4\%}$} & 0.998\textcolor{Green}{$\downarrow_{4.3\%}$} & 1.178\textcolor{Green}{$\downarrow_{4.6\%}$} \\
\bottomrule
\end{tabular}
\end{table}
\subsection{Taking UniTalker as a Foundation Model}
\noindent\textbf{Fine-tuning UniTalker on Seen Annotations.}
UniTalker is motivated to improve the overall performance and needs to consider the trade-off in performance across different datasets.
To get consistent improvement on every dataset, we fine-tune UniTalker on each individual dataset and get eight fine-tuned models, denoted as L-FT. As evidenced by~\cref{tab:LVE_single_all_and_finetune}, this fine-tuning process further enhances performance on every dataset. Compared with L-[D*], L-FT leads to better precision across all datasets, including the hard-case datasets like D4 with emotional speeches ~\cite{livingstone2018ryerson} and D7 with songs. The largest two LVE reductions are 11.9\% on D1 and 10.8\% on D0. The average LVE drop across datasets is 6.3\%. 

\noindent\textbf{Fine-tuning UniTalker on Unseen Annotations.}
\label{sec:Results-Annotation-Transfer}
We train UniTalker-[D1-D7] and fine-tune it on D0 (BIWI). As a comparison, we directly fine-tune Wav2vec2-xlsr-53~\cite{Wav2Vec2_XLSR_53} on D0. When fine-tuning UniTalker-[D1-D7], we only keep the weights of UniTalker encoder and reinitialize the weights of decoder, to ensure fair comparison. 
The original D0 training set contains 190 sequences, with 32 utterances for each speaker and 2 utterances missing. We iteratively discard half of the training set, leaving 96, 48, 24, 12 and 6 sequences. The smallest subset contains only one utterance per speaker, and the utterance content is identical across all speakers. We fine-tune UniTalker-[D1-D7] and Wav2vec2-xlsr-53 on D0 and each subset. 
Fig.~\ref{fig:miniBIWI_transfer_compare} shows that fine-tuning UniTalker-[D1-D7] always yields better precision. It requires less than half of the data to get comparable performance. Moreover, fine-tuning UniTalker on D0-half, achieves lower LVE, \ie, 4.197$\times10^{-4}$ than that of previous state-of-the-art model~\cite{peng2023selftalk} trained on D0-full, \ie, 4.249$\times10^{-4}$.
\section{Ablation Study}
To analyse the effects of the different components of UniTalker, we conducted ablation studies in terms of audio encoder, motion decoder and the frequency adaptor. Please refer to Supplementary Materials for the latter two.

\noindent\textbf{Effect of Pre-trained Audio Encoder.}
\label{sec:audio encoder ablation}
Bao \etal \cite{bao2023learning} shows that the self-supervised pre-trained audio features substantially boost the performance for audio-driven facial animation, compared with handcrafted features.
Based on this observation, we investigate the effect of different pre-trained audio encoders. 
Wav2vec2-base-960h~\cite{Wav2Vec2_Base_960h,baevski2020wav2vec} is pre-trained on 960 hours of English speech. 
Wavlm-base~\cite{WavLM_Base} is pre-trained on the same dataset with different pre-training method.
Wavlm-base-plus~\cite{WavLM_Base_Plus} has the same model size with Wav2vec2-base-960h and Wavlm-base, but is pre-trained on 94k hours of audios in 23 languages. 
Wav2vec2-xlsr-53~\cite{Wav2Vec2_XLSR_53,conneau2020unsupervised} is a larger audio encoder and pre-trained on 56k hours of audios in 53 languages.
We train UniTalker on A2F-Bench, based on these four audio encoders and report LVE on each test set.
As shown in \cref{table:effect_of_audio_encoder}, UniTalker based on Wav2vec2-base-960h shows suboptimal performance. 
Wavlm-base shows significant improvement over Wav2vec2-base-960h due to better pre-training method. 
With scaled-up pre-training data, Wavlm-base-plus shows better performance over Wavlm-base. 
Benifit from the diversity of pre-training data and larger capacity, Wav2vec2-xlsr-53 leads to an overall performance improvement.
\cref{table:effect_of_audio_encoder} shows that the downstream UniTalker precision is largely affected by the pre-trained audio encoder from three aspects, including the pre-training method, the scale and diversity of pre-training dataset and the capacity of pre-training backbone.
\begin{table}[tb]
\caption{The effect of pre-trained audio encoders.
The first row indicates the test dataset. LVE is in ${10^{-4}}$ for D0, ${10^{-6}}$ ${m^{2}}$ for D1-D3 and ${10^{-5}}$ ${m^{2}}$ for D4-D7.}
\centering 
\label{table:effect_of_audio_encoder} 
\scriptsize 
\begin{tabular}{l | c  c  c  c  c  c  c  c  }
\toprule
\textbf{Audio Encoder} &  \textbf{D0}  & \textbf{D1}  & \textbf{D2} & \textbf{D3} & \textbf{D4} & \textbf{D5}  & \textbf{D6} &\textbf{D7} \\
\midrule
Wav2Vec2-Base-960h~\cite{Wav2Vec2_Base_960h} & 4.491 & 9.916 & 9.887 & 9.812 & 1.585 & 2.217 & 1.351 & 1.409 \\
WavLM-Base~\cite{chen2022wavlm} & 4.033 & 8.269 & 9.253 & 9.117 & 1.417 & 2.044 & 1.184 & 1.340 \\
WavLM-Base-Plus~\cite{WavLM_Base_Plus}       & 4.080 & \textbf{8.136} & 9.776 & 9.053 & 1.392 & \textbf{1.975} & 1.158 & 1.264 \\
Wav2Vec-XLSR-53~\cite{Wav2Vec2_XLSR_53}      & \textbf{3.859} & 8.303 & \textbf{8.648} & 
 \textbf{8.991} & \textbf{1.326} & 2.056 & \textbf{1.145} & \textbf{1.211} \\
\bottomrule 
\end{tabular} 
\end{table}

\section{Conclusion and Discussion}

We propose UniTalker, which effectively exploits the existing datasets with inconsistent annotation format.
The model precision benefits from the increased scale and diversity of A2F-Bench. 
%
The experiment shows that the pre-trained UniTalker has the potential to serve as a foundation model for more audio-to-face tasks, especially when the data is scarce.
%

\noindent\textbf{Limitations and Future Works.} \cref{tab:LVE_single_all_and_finetune} indicates that UniTalker shows better precision on most datasets than the corresponding individual models. However, achieving consistent improvement over every dataset requires dataset-specific fine-tuning.
%
The potential for enhancing model capacity to alleviate performance trade-offs across diverse datasets remains an open problem.
%
%
Meanwhile, \cref{fig:miniBIWI_transfer_compare} indicates that the pre-trained UniTalker exhibits promise as the foundation model for audio-driven facial animation tasks. 
Nonetheless, the data scale used for UniTalker, \ie, 18.53 hours, is still considerably smaller than that used for training the audio encoder, \ie., 56k hours. 
%
%
Exploring the utilization of large-scale datasets with suboptimal data quality, such as BEAT and Talkshow, represents a promising future direction. Applying UniTalker to 2D facial animation~\cite{tian2024emo,xu2024vasa,qiu2024relitalk} to enhance consistency under large head poses is also a worthwhile pursuit.
%





%
%
\bibliographystyle{splncs04}
\bibliography{main}

\begin{thebibliography}{10}
\providecommand{\url}[1]{\texttt{#1}}
\providecommand{\urlprefix}{URL }
\providecommand{\doi}[1]{https://doi.org/#1}

\bibitem{openai_tts}
{OpenAI Text-to-Speech}. \url{https://platform.openai.com/docs/guides/text-to-speech/}

\bibitem{amberg2007optimal}
Amberg, B., Romdhani, S., Vetter, T.: Optimal step nonrigid icp algorithms for surface registration. In: 2007 IEEE conference on computer vision and pattern recognition. pp.~1--8. IEEE (2007)

\bibitem{anyi2023dynamic}
Anyi, R., Xuekun, J., Yuwei, G., Linning, X., Lei, Y., Libiao, J., Dahua, L., Bo, D.: Dynamic storyboard generation in an engine-based virtual environment for video production. arXiv preprint arXiv:2301.12688  (2023)

\bibitem{Wav2Vec2_Base_960h}
Baevski, A., Zhou, Y., Mohamed, A., Auli, M.: {Wav2Vec2-Base-960h}. \url{https://huggingface.co/facebook/wav2vec2-base-960h}

\bibitem{baevski2020wav2vec}
Baevski, A., Zhou, Y., Mohamed, A., Auli, M.: wav2vec 2.0: A framework for self-supervised learning of speech representations. Advances in neural information processing systems  \textbf{33},  12449--12460 (2020)

\bibitem{bao2023learning}
Bao, L., Zhang, H., Qian, Y., Xue, T., Chen, C., Zhe, X., Kang, D.: Learning audio-driven viseme dynamics for 3d face animation. arXiv preprint arXiv:2301.06059  (2023)

\bibitem{black2023bedlam}
Black, M.J., Patel, P., Tesch, J., Yang, J.: Bedlam: A synthetic dataset of bodies exhibiting detailed lifelike animated motion. In: Proceedings of the IEEE/CVF Conference on Computer Vision and Pattern Recognition. pp. 8726--8737 (2023)

\bibitem{bolkart2023instant}
Bolkart, T., Li, T., Black, M.J.: Instant multi-view head capture through learnable registration. In: Proceedings of the IEEE/CVF Conference on Computer Vision and Pattern Recognition. pp. 768--779 (2023)

\bibitem{cai2023digital}
Cai, Z., Jiang, J., Qing, Z., Guo, X., Zhang, M., Lin, Z., Mei, H., Wei, C., Wang, R., Yin, W., et~al.: Digital life project: Autonomous 3d characters with social intelligence. arXiv preprint arXiv:2312.04547  (2023)

\bibitem{cai2023smpler}
Cai, Z., Yin, W., Zeng, A., Wei, C., Sun, Q., Yanjun, W., Pang, H.E., Mei, H., Zhang, M., Zhang, L., Loy, C.C., Yang, L., Liu, Z.: Smpler-x: Scaling up expressive human pose and shape estimation. In: Oh, A., Neumann, T., Globerson, A., Saenko, K., Hardt, M., Levine, S. (eds.) Advances in Neural Information Processing Systems. vol.~36, pp. 11454--11468. Curran Associates, Inc. (2023)

\bibitem{chai2023hiface}
Chai, Z., Zhang, T., He, T., Tan, X., Baltrusaitis, T., Wu, H., Li, R., Zhao, S., Yuan, C., Bian, J.: Hiface: High-fidelity 3d face reconstruction by learning static and dynamic details. In: Proceedings of the IEEE/CVF International Conference on Computer Vision. pp. 9087--9098 (2023)

\bibitem{chen2023understanding}
Chen, H., Wang, J., Shah, A., Tao, R., Wei, H., Xie, X., Sugiyama, M., Raj, B.: Understanding and mitigating the label noise in pre-training on downstream tasks. arXiv preprint arXiv:2309.17002  (2023)

\bibitem{WavLM_Base}
Chen, S., Wang, C., Chen, Z., Wu, Y., Liu, S., Chen, Z., Li, J., Kanda, N., Yoshioka, T., Xiao, X., et~al.: {WavLM-Base}. \url{https://huggingface.co/microsoft/wavlm-base}

\bibitem{WavLM_Base_Plus}
Chen, S., Wang, C., Chen, Z., Wu, Y., Liu, S., Chen, Z., Li, J., Kanda, N., Yoshioka, T., Xiao, X., et~al.: {WavLM-Base-Plus}. \url{https://huggingface.co/microsoft/wavlm-base-plus}

\bibitem{chen2022wavlm}
Chen, S., Wang, C., Chen, Z., Wu, Y., Liu, S., Chen, Z., Li, J., Kanda, N., Yoshioka, T., Xiao, X., et~al.: Wavlm: Large-scale self-supervised pre-training for full stack speech processing. IEEE Journal of Selected Topics in Signal Processing  \textbf{16}(6),  1505--1518 (2022)

\bibitem{Wav2Vec2_XLSR_53}
Conneau, A., Baevski, A., Collobert, R., Mohamed, A., Auli, M.: {Wav2Vec2-XLSR-53}. \url{https://huggingface.co/facebook/wav2vec2-large-xlsr-53}

\bibitem{conneau2020unsupervised}
Conneau, A., Baevski, A., Collobert, R., Mohamed, A., Auli, M.: Unsupervised cross-lingual representation learning for speech recognition. arXiv preprint arXiv:2006.13979  (2020)

\bibitem{xrfeitoria}
Contributors, X.: Openxrlab synthetic data rendering toolbox. \url{https://github.com/openxrlab/xrfeitoria} (2023)

\bibitem{cudeiro2019capture}
Cudeiro, D., Bolkart, T., Laidlaw, C., Ranjan, A., Black, M.J.: Capture, learning, and synthesis of 3d speaking styles. In: Proceedings of the IEEE/CVF Conference on Computer Vision and Pattern Recognition. pp. 10101--10111 (2019)

\bibitem{fan2022faceformer}
Fan, Y., Lin, Z., Saito, J., Wang, W., Komura, T.: Faceformer: Speech-driven 3d facial animation with transformers. In: Proceedings of the IEEE/CVF Conference on Computer Vision and Pattern Recognition. pp. 18770--18780 (2022)

\bibitem{fanelli20103}
Fanelli, G., Gall, J., Romsdorfer, H., Weise, T., Van~Gool, L.: A 3-d audio-visual corpus of affective communication. IEEE Transactions on Multimedia  \textbf{12}(6),  591--598 (2010)

\bibitem{filntisis2022visual}
Filntisis, P.P., Retsinas, G., Paraperas-Papantoniou, F., Katsamanis, A., Roussos, A., Maragos, P.: Visual speech-aware perceptual 3d facial expression reconstruction from videos. arXiv preprint arXiv:2207.11094  (2022)

\bibitem{grosman2021xlsr53-large-english}
Grosman, J.: Fine-tuned {XLSR}-53 large model for speech recognition in {E}nglish. \url{https://huggingface.co/jonatasgrosman/wav2vec2-large-xlsr-53-english} (2021)

\bibitem{ho2022classifier}
Ho, J., Salimans, T.: Classifier-free diffusion guidance. arXiv preprint arXiv:2207.12598  (2022)

\bibitem{facebook/hubert-base-ls960}
Hsu, W.N., Bolte, B., Tsai, Y.H.H., Lakhotia, K., Salakhutdinov, R., Mohamed, A.: facebook/hubert-base-ls960. \url{https://huggingface.co/facebook/hubert-base-ls960}

\bibitem{hsu2021hubert}
Hsu, W.N., Bolte, B., Tsai, Y.H.H., Lakhotia, K., Salakhutdinov, R., Mohamed, A.: Hubert: Self-supervised speech representation learning by masked prediction of hidden units. IEEE/ACM Transactions on Audio, Speech, and Language Processing  \textbf{29},  3451--3460 (2021)

\bibitem{iwase2020song2face}
Iwase, S., Kato, T., Yamaguchi, S., Yukitaka, T., Morishima, S.: Song2face: Synthesizing singing facial animation from audio. In: SIGGRAPH Asia 2020 Technical Communications, pp.~1--4 (2020)

\bibitem{karras2017audio}
Karras, T., Aila, T., Laine, S., Herva, A., Lehtinen, J.: Audio-driven facial animation by joint end-to-end learning of pose and emotion. ACM Transactions on Graphics (TOG)  \textbf{36}(4),  1--12 (2017)

\bibitem{li2017learning}
Li, T., Bolkart, T., Black, M.J., Li, H., Romero, J.: Learning a model of facial shape and expression from 4d scans. ACM Trans. Graph.  \textbf{36}(6),  194--1 (2017)

\bibitem{lin2022high}
Lin, Z., Lin, J., Li, L., Yuan, Y., Zou, Z.: High-quality 3d face reconstruction with affine convolutional networks. In: Proceedings of the 30th ACM International Conference on Multimedia. pp. 2495--2503 (2022)

\bibitem{liu2023emage}
Liu, H., Zhu, Z., Becherini, G., Peng, Y., Su, M., Zhou, Y., Iwamoto, N., Zheng, B., Black, M.J.: Emage: Towards unified holistic co-speech gesture generation via masked audio gesture modeling. arXiv preprint arXiv:2401.00374  (2023)

\bibitem{liu2022beat}
Liu, H., Zhu, Z., Iwamoto, N., Peng, Y., Li, Z., Zhou, Y., Bozkurt, E., Zheng, B.: Beat: A large-scale semantic and emotional multi-modal dataset for conversational gestures synthesis. In: European Conference on Computer Vision. pp. 612--630. Springer (2022)

\bibitem{livingstone2018ryerson}
Livingstone, S.R., Russo, F.A.: The ryerson audio-visual database of emotional speech and song (ravdess): A dynamic, multimodal set of facial and vocal expressions in north american english. PloS one  \textbf{13}(5),  e0196391 (2018)

\bibitem{dad3dheads}
Martyniuk, T., Kupyn, O., Kurlyak, Y., Krashenyi, I., Matas, J., Sharmanska, V.: Dad-3dheads: A large-scale dense, accurate and diverse dataset for 3d head alignment from a single image. In: Proc. IEEE Conf. on Computer Vision and Pattern Recognition (CVPR) (2022)

\bibitem{pan2023renderme}
Pan, D., Zhuo, L., Piao, J., Luo, H., Cheng, W., Yuxin, W., Fan, S., Liu, S., Yang, L., Dai, B., et~al.: Renderme-360: A large digital asset library and benchmarks towards high-fidelity head avatars. In: Thirty-seventh Conference on Neural Information Processing Systems Datasets and Benchmarks Track (2023)

\bibitem{peng2023selftalk}
Peng, Z., Luo, Y., Shi, Y., Xu, H., Zhu, X., Liu, H., He, J., Fan, Z.: Selftalk: A self-supervised commutative training diagram to comprehend 3d talking faces. arXiv preprint arXiv:2306.10799  (2023)

\bibitem{peng2023emotalk}
Peng, Z., Wu, H., Song, Z., Xu, H., Zhu, X., He, J., Liu, H., Fan, Z.: Emotalk: Speech-driven emotional disentanglement for 3d face animation. In: Proceedings of the IEEE/CVF International Conference on Computer Vision. pp. 20687--20697 (2023)

\bibitem{qing2023story}
Qing, Z., Cai, Z., Yang, Z., Yang, L.: Story-to-motion: Synthesizing infinite and controllable character animation from long text. In: SIGGRAPH Asia 2023 Technical Communications, pp.~1--4 (2023)

\bibitem{qiu2024relitalk}
Qiu, H., Chen, Z., Jiang, Y., Zhou, H., Fan, X., Yang, L., Wu, W., Liu, Z.: Relitalk: Relightable talking portrait generation from a single video. International Journal of Computer Vision pp. 1--16 (2024)

\bibitem{richard2021meshtalk}
Richard, A., Zollh{\"o}fer, M., Wen, Y., De~la Torre, F., Sheikh, Y.: Meshtalk: 3d face animation from speech using cross-modality disentanglement. In: Proceedings of the IEEE/CVF International Conference on Computer Vision. pp. 1173--1182 (2021)

\bibitem{ross2008incremental}
Ross, D.A., Lim, J., Lin, R.S., Yang, M.H.: Incremental learning for robust visual tracking. International journal of computer vision  \textbf{77},  125--141 (2008)

\bibitem{rossler1901learning}
Rossler, A., Cozzolino, D., Verdoliva, L., Verdoliva, L., Riess, C., Thies, J., Nie{\ss}ner, M.F.: Learning to detect manipulated facial images. arxiv 2019. arXiv preprint arXiv:1901.08971

\bibitem{shimba2015talking}
Shimba, T., Sakurai, R., Yamazoe, H., Lee, J.H.: Talking heads synthesis from audio with deep neural networks. In: 2015 IEEE/SICE International Symposium on System Integration (SII). pp. 100--105. IEEE (2015)

\bibitem{siyao2023duolando}
Siyao, L., Gu, T., Yang, Z., Lin, Z., Liu, Z., Ding, H., Yang, L., Loy, C.C.: Duolando: Follower gpt with off-policy reinforcement learning for dance accompaniment. In: The Twelfth International Conference on Learning Representations (2023)

\bibitem{stan2023facediffuser}
Stan, S., Haque, K.I., Yumak, Z.: Facediffuser: Speech-driven 3d facial animation synthesis using diffusion. In: Proceedings of the 16th ACM SIGGRAPH Conference on Motion, Interaction and Games. pp. 1--11 (2023)

\bibitem{sun2024aios}
Sun, Q., Wang, Y., Zeng, A., Yin, W., Wei, C., Wang, W., Mei, H., Leung, C.S., Liu, Z., Yang, L., et~al.: Aios: All-in-one-stage expressive human pose and shape estimation. In: Proceedings of the IEEE/CVF Conference on Computer Vision and Pattern Recognition. pp. 1834--1843 (2024)

\bibitem{suwajanakorn2017synthesizing}
Suwajanakorn, S., Seitz, S.M., Kemelmacher-Shlizerman, I.: Synthesizing obama: learning lip sync from audio. ACM Transactions on Graphics (ToG)  \textbf{36}(4),  1--13 (2017)

\bibitem{tian2024emo}
Tian, L., Wang, Q., Zhang, B., Bo, L.: Emo: Emote portrait alive-generating expressive portrait videos with audio2video diffusion model under weak conditions. arXiv preprint arXiv:2402.17485  (2024)

\bibitem{veit2017learning}
Veit, A., Alldrin, N., Chechik, G., Krasin, I., Gupta, A., Belongie, S.: Learning from noisy large-scale datasets with minimal supervision. In: Proceedings of the IEEE conference on computer vision and pattern recognition. pp. 839--847 (2017)

\bibitem{wang2011text}
Wang, L., Han, W., Soong, F.K., Huo, Q.: Text driven 3d photo-realistic talking head. In: Twelfth Annual Conference of the International Speech Communication Association (2011)

\bibitem{wang2023zolly}
Wang, W., Ge, Y., Mei, H., Cai, Z., Sun, Q., Wang, Y., Shen, C., Yang, L., Komura, T.: Zolly: Zoom focal length correctly for perspective-distorted human mesh reconstruction. In: Proceedings of the IEEE/CVF International Conference on Computer Vision. pp. 3925--3935 (2023)

\bibitem{wu2023mmface4d}
Wu, H., Jia, J., Xing, J., Xu, H., Wang, X., Wang, J.: Mmface4d: A large-scale multi-modal 4d face dataset for audio-driven 3d face animation. arXiv preprint arXiv:2303.09797  (2023)

\bibitem{wu2023speech}
Wu, H., Zhou, S., Jia, J., Xing, J., Wen, Q., Wen, X.: Speech-driven 3d face animation with composite and regional facial movements. In: Proceedings of the 31st ACM International Conference on Multimedia. pp. 6822--6830 (2023)

\bibitem{wuu2022multiface}
Wuu, C.h., Zheng, N., Ardisson, S., Bali, R., Belko, D., Brockmeyer, E., Evans, L., Godisart, T., Ha, H., Huang, X., et~al.: Multiface: A dataset for neural face rendering. arXiv preprint arXiv:2207.11243  (2022)

\bibitem{xing2023codetalker}
Xing, J., Xia, M., Zhang, Y., Cun, X., Wang, J., Wong, T.T.: Codetalker: Speech-driven 3d facial animation with discrete motion prior. In: Proceedings of the IEEE/CVF Conference on Computer Vision and Pattern Recognition. pp. 12780--12790 (2023)

\bibitem{xu2024vasa}
Xu, S., Chen, G., Guo, Y.X., Yang, J., Li, C., Zang, Z., Zhang, Y., Tong, X., Guo, B.: Vasa-1: Lifelike audio-driven talking faces generated in real time. arXiv preprint arXiv:2404.10667  (2024)

\bibitem{yang2020learn}
Yang, L., Huang, Q., Huang, H., Xu, L., Lin, D.: Learn to propagate reliably on noisy affinity graphs. In: European Conference on Computer Vision. pp. 447--464. Springer (2020)

\bibitem{yang2023synbody}
Yang, Z., Cai, Z., Mei, H., Liu, S., Chen, Z., Xiao, W., Wei, Y., Qing, Z., Wei, C., Dai, B., Wu, W., Qian, C., Lin, D., Liu, Z., Yang, L.: Synbody: Synthetic dataset with layered human models for 3d human perception and modeling. In: Proceedings of the IEEE/CVF International Conference on Computer Vision (ICCV). pp. 20282--20292 (October 2023)

\bibitem{yi2023generating}
Yi, H., Liang, H., Liu, Y., Cao, Q., Wen, Y., Bolkart, T., Tao, D., Black, M.J.: Generating holistic 3d human motion from speech. In: Proceedings of the IEEE/CVF Conference on Computer Vision and Pattern Recognition. pp. 469--480 (2023)

\bibitem{yin2024whac}
Yin, W., Cai, Z., Wang, R., Wang, F., Wei, C., Mei, H., Xiao, W., Yang, Z., Sun, Q., Yamashita, A., et~al.: Whac: World-grounded humans and cameras. arXiv preprint arXiv:2403.12959  (2024)

\bibitem{zeng2022smoothnet}
Zeng, A., Yang, L., Ju, X., Li, J., Wang, J., Xu, Q.: Smoothnet: A plug-and-play network for refining human poses in videos. In: European Conference on Computer Vision. pp. 625--642. Springer (2022)

\bibitem{zeng20233d}
Zeng, L., Chen, L., Bao, W., Li, Z., Xu, Y., Yuan, J., Kalantari, N.K.: 3d-aware facial landmark detection via multi-view consistent training on synthetic data. In: Proceedings of the IEEE/CVF Conference on Computer Vision and Pattern Recognition. pp. 12747--12758 (2023)

\bibitem{zhang2024large}
Zhang, M., Jin, D., Gu, C., Hong, F., Cai, Z., Huang, J., Zhang, C., Guo, X., Yang, L., He, Y., et~al.: Large motion model for unified multi-modal motion generation. arXiv preprint arXiv:2404.01284  (2024)

\bibitem{zhang2021flow}
Zhang, Z., Li, L., Ding, Y., Fan, C.: Flow-guided one-shot talking face generation with a high-resolution audio-visual dataset. In: Proceedings of the IEEE/CVF Conference on Computer Vision and Pattern Recognition. pp. 3661--3670 (2021)

\bibitem{zhao2024media2face}
Zhao, Q., Long, P., Zhang, Q., Qin, D., Liang, H., Zhang, L., Zhang, Y., Yu, J., Xu, L.: Media2face: Co-speech facial animation generation with multi-modality guidance. arXiv preprint arXiv:2401.15687  (2024)

\end{thebibliography}

\title{UniTalker: Scaling up Audio-Driven 3D Facial Animation through A Unified Model} 
\subtitle{-- Supplementary Materials --}

\titlerunning{UniTalker}

\author{Xiangyu Fan\inst{}\orcidlink{0000-0002-3446-524X} \and
Jiaqi Li\inst{}\orcidlink{0000-0002-0058-0266} \and
Zhiqian Lin\inst{}\orcidlink{0009-0005-5971-1928} \and
Weiye Xiao\inst{}\orcidlink{0000-0003-3015-3609} \and
Lei Yang\inst{}\orcidlink{0000-0002-0571-5924}\textsuperscript{\Letter}}
\authorrunning{X.~Fan et al.}

\institute{SenseTime Research, China \\
\email{\{fanxiangyu, lijiaqi2, linzhiqian, xiaoweiye1, yanglei\}@sensetime.com} 
}

\maketitle

\section{Demonstration Video}
We present a brief demonstration of UniTalker in the attached video and the project page\footnote{Homepage: \url{https://github.com/X-niper/UniTalker}}.
Our model exhibits the ability to generate realistic facial motion with different audio inputs, including clean and noisy voices in various languages, text-to-speech-generated audios, and even noisy songs accompanied by background music.
Notably, our model excels at predicting face emotion according to the input audio. 
Although Vocaset consists of only neutral voices and emotionless facial motion, our model effectively infuse emotional facial motion into the generated faces.
In contrast, previous models~\cite{fan2022faceformer,xing2023codetalker,peng2023selftalk,stan2023facediffuser} trained exclusively on Vocaset struggle to generate facial motion beyond neutral expression, even when the input audio carries strong emotional cues.
When given audios with strong emotion, the generated faces may exhibit over-exaggerated and unnatural emotion for Vocaset annotation, since Vocaset contains only neutral emotion (see Main Paper Fig. 6a). The model needs to "guess" and generate out-of-domain facial motion for Vocaset annotation.
We include this failure case in the attached video.
The synthesized audio mentioned in the demo video is generated using OpenAI's Text-to-Speech voice~\cite{openai_tts}.
\section{Additional Experiments}

\subsection{Comparison between TCN and Transformer}
We do experiments on TCN and transformer for the motion decoder. Like~\cite{bao2023learning}, the transformer refers to the non-autoregressive transformer encoder architecture. Both TCN and transformer have 256 channels and 3 layers. The transformer is with 4 heads. \cref{supp_tab:tcn_and_transformer} shows that TCN leads to lower LVE on most datasets.
\begin{table}[tb]
\small
\centering
\caption{Effect of motion decoder architecture. UniTalker adopts Temporal Convolutional Network (TCN) due to the lower LVE. Transformer denotes the non-autoregressive transformer encoder architecture.
LVE is in ${10^{-4}}$ for D0, ${10^{-6}}$ ${m^{2}}$ for D1-D3 and ${10^{-5}}$ ${m^{2}}$ for D4-D7.
}
\label{supp_tab:tcn_and_transformer}
\begin{tabular}{c | c c c c c c c c}
\toprule
\textbf{Method} & \textbf{D0}  & \textbf{D1}  & \textbf{D2} & \textbf{D3} & \textbf{D4} & \textbf{D5}  & \textbf{D6} &\textbf{D7} \\

\midrule
\textbf{TCN} &          \textbf{3.859} & \textbf{8.303} & \textbf{8.648} & \textbf{8.991} & \textbf{1.326} & 2.056 & 1.145 & \textbf{1.211} \\
\textbf{Transformer} &  3.971 & 8.679 & 8.756 & 9.567 & 1.335 & \textbf{1.993} & \textbf{1.092} & 1.372 \\

\bottomrule
\end{tabular}
\end{table}

\subsection{One-shot Learning}

To explore the possibility of one-shot tuning for the pre-trained UniTalker model, we conduct an experiment using a single audio-visual pair from Vocaset as the training set. The remaining audio-visual pairs from the same speaker are allocated to the validation set, which consists of 38 pairs. Initially, we train the UniTalker model on the combined datasets [D0, D2-D7], and then perform fine-tuning on this model using the one-sentence training set. We compare two different tuning methods: (a) tuning all parameters except for the parameters in the TCN of Wav2vec2-xlsr-53~\cite{Wav2Vec2_XLSR_53}, and (b) tuning only the decoder part while freezing the weights of the audio encoder.
We also include a control group where the model is tuned directly from Wav2vec2-xlsr-53 on the one-sentence training set. The best validation Lip Vertex Error (LVE) and Lip Vertex Distance (LVD) for each method are listed in \cref{supp_tab:one_shot_training}.
LVD is computed as the average over all frames of maximal Euclidean distance of the lip vertices to the ground truth. 
The results demonstrate that in the one-shot training scenario, fine-tuning the decoder part helps to prevent over-fitting and leads to better precision. As demonstrated in the attached video, the decoder-tuned model achieves visually pleasant results, while the directly trained model results in twitching mouth motion. 
\begin{table}[tb]
\centering
\caption{
Results on one-shot training experiments.
The one-shot training is conducted by fine-tuning three models: (1) Wav2vec2-xlsr-53, (2) UniTalker-L-[D0, D2-D7], and (3) the decoder component of UniTalker-L-[D0, D2-D7]. These models were fine-tuned using a one-utterance subset of Vocaset. We then evaluated the models on a test set comprising 38 utterances from the same speaker, reporting both LVE and LVD metrics.
}
\label{supp_tab:one_shot_training}
\small
\begin{tabular}{c | c c }
\toprule
\multirow{2}{*}{ \textbf{Method} } & \textbf{ LVE } & \textbf{ LVD } 
 \\
 & ${ \times {10^{-5} m^2}}$ & $mm$ \\
\midrule
\textbf{Wav2vec2-xlsr-53}       & 3.4812   &  5.1479   \\
\textbf{UniTalker-L-[D0, D2-D7]}         & 2.2169   &  4.2043   \\
\textbf{Decoder of UniTalker-L-[D0, D2-D7]}   & \textbf{2.2070}   &  \textbf{4.1614}   \\
\bottomrule
\end{tabular}
\end{table}

\subsection{Effect of Frequency Adaptor Position}

The effect of frequency adaptor position is presented in \cref{supp_tab:effect_of_frequency_adaptor}. We find that placing the frequency adaptor behind the transformer of the audio encoder yields higher precision, compared with the original position in FaceFormer~\cite{fan2022faceformer} and CodeTalker~\cite{xing2023codetalker}. In the UniTalker model, the transformer within the audio encoder receives audio features at a consistent frequency, while the TCN decoder body receives contextualized audio features with varying frequencies, which can be viewed as a scaling augmentation. It is hypothesized that this augmentation contributes to improved generalization. 
\begin{table}[tb]
\centering
\caption{\textbf{Effect of frequency adaptor position.} Pos-0 means that the frequency adaptor is placed between the TCN and transformer in the audio encoder. Pos-1 means that the frequency adaptor is placed behind the transformer. The metric is LVE.
LVE is in ${10^{-4}}$ for D0, ${10^{-6}}$ ${m^{2}}$ for D1-D3 and ${10^{-5}}$ ${m^{2}}$ for D4-D7.
} 
\label{supp_tab:effect_of_frequency_adaptor}
\small
\begin{tabular}{l | c c c c c c c c}
\toprule
\textbf{Adaptor Position}  &  \textbf{D0} & \textbf{D1} & \textbf{D2} & \textbf{D3} & \textbf{D4} & \textbf{D5} & \textbf{D6} & \textbf{D7} \\
\midrule
\textbf{Pos-0}  &    3.951 &        \textbf{8.118} &          8.808 &           9.201 &             1.408 &               2.131 &       \textbf{1.096} &             1.289  \\
\textbf{Pos-1 (UniTalker)} &    \textbf{3.859} &        8.303 &          \textbf{8.648} &           \textbf{8.991} &             \textbf{1.326} &               \textbf{2.056} &       1.145 &             \textbf{1.211}  \\
\bottomrule
\end{tabular}
\end{table}

\subsection{Comparison between Regressive and Autoregressive Decoder}

We explore the effect of the extensively adopted autoregression in prior works like FaceFormer~\cite{fan2022faceformer} and CodeTalker~\cite{xing2023codetalker}. We do experiments with the official FaceFormer~\cite{fan2022faceformer} codebase on BIWI and Vocaset. We remove autoregression by ignoring the previously predicted face vertices displacement. As listed in \cref{supp_tab:auto_regression}, removing autoregression from FaceFormer model doesn't degrade the accuracy but improves the inference speed to nearly 30 times. We also examine the visualization and find no remarkable differences between the original and the modified model. We adopt the non-autoregressive decoder due to the improvded inference speed, as in prior works like TalkShow~\cite{yi2023generating} and SelfTalk~\cite{peng2023selftalk}.
\begin{table}[tb]
\centering
\caption{Effect of autoregression. Removing autoregression from FaceFormer does not degrade the precision but improves the inference speed to nearly 30 times.}
\label{supp_tab:auto_regression}
\small
\begin{tabular}{c | c c | c c}
\toprule
\multirow{3}{*}{\textbf{Autoregression}} & \multicolumn{2}{c|}{BIWI-Test-A} & \multicolumn{2}{c}{VOCA-Test} \\ & \textbf{LVE}  $\downarrow$ & \textbf{Time} $\downarrow$ & \textbf{LVE} $\downarrow$   & \textbf{Time} $\downarrow$ \\

 & ($\times 10^{-4}$) & ($s$) & ($\times 10^{-5} m^2$) & ($s$)\\
\midrule
\cmark     &  4.9836  &  0.705    & 1.1221  &  0.624  \\
\xmark &  4.9259   &  0.024   & 1.1453 &  0.021  \\
\bottomrule
\end{tabular}
\end{table}

\subsection{The Importance of Hard-case Datasets}
We examine the importance of hard-case dataset, \ie, RAVDESS and our collected multilingual song dataset. RAVDESS contains audios with eight kinds of emotions. Our collected multilingual song dataset contains audios with different kinds of song styles. 
We train UniTalker-[D0-D3] and UniTalker-[D0-D4] and test them on D4 test set, conditioning on the pivot identity for fair comparison. 
Since D3 and D4 share the same annotation type \cref{subfig:decoder_details}, this evaluation can be done directly without further tuning. 
Similarly, we train UniTalker-[D0-D6] and UniTalker-[D0-D7], and subsequently evaluate their performance on D7 test set, conditioning on the pivot identity. \cref{supp_tab:hard_case_dataset_is_necessary} shows that the models trained on datasets lacking strong emotions or songs struggle to handle these challenging cases effectively.
The inability to adequately handle such cases suggests the necessity of incorporating datasets containing strong emotional and musical content during training. 
By including these diverse and challenging scenarios in the training data, the model can learn to better handle similar cases during inference, thereby improving overall performance.
\begin{table}[tb]
\centering
\caption{The comparison of LVD (${mm}$) between models trained on datasets with and without hard-case data. The LVD differences between L-[D0-D3] and L-[D0-D4] on D4 test set shows the contribution of data with strong emotion. The LVD differences between L-[D0-D6] and L-[D0-D7] on D7 test set shows the contribution of multilingual songs. In this experiment, all inferences are conditioned on the pivot identity for fair comparison.}
\label{supp_tab:hard_case_dataset_is_necessary}
\small
\begin{tabular}{l | c c}
\toprule
\textbf{Model} & \textbf{D4} & \textbf{D7} \\
\midrule
\textbf{UniTalker-[D0-D3]}  & 5.5730 & - \\
\textbf{UniTalker-[D0-D4]}  & \textbf{3.6875} & - \\
\textbf{UniTalker-[D0-D6]}  & - & 6.4963 \\
\textbf{UniTalker-[D0-D7]}  & - & \textbf{3.0958} \\
\bottomrule
\end{tabular}
\end{table}

\section{Detailed Implementation}

\subsection{Vanilla Multi-Head Model and UniTalker Model}
\begin{figure}[tb]
  \centering
  \begin{subfigure}{0.7\linewidth}
    \includegraphics[width=\linewidth]{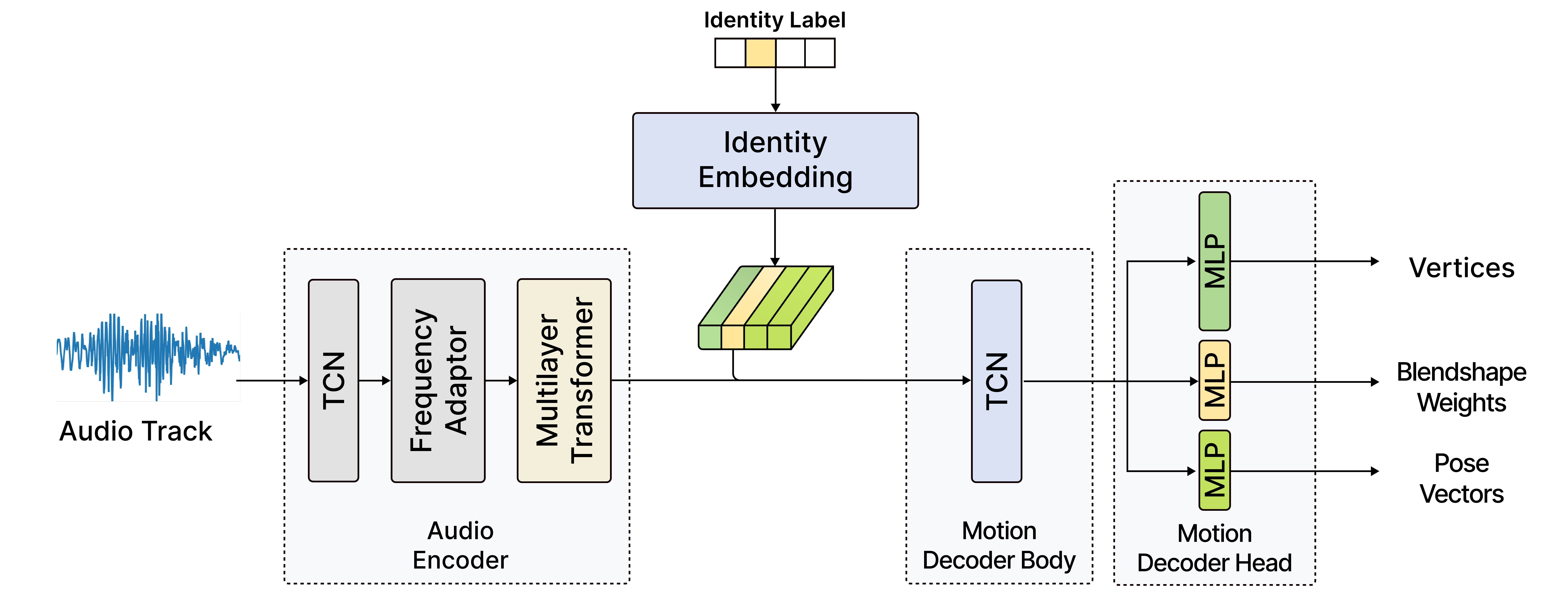}
    \caption{Vanilla Multi-Head Model}
    \label{subfig:unitalker_archi_vanilla}
  \end{subfigure}
  \begin{subfigure}{0.7\linewidth}
    \includegraphics[width=\linewidth]{figures/images/unitalker_archi_unitalker_v2.pdf}
    \caption{UniTalker Model}
    \label{subfig:unitalker_archi_unitalker}
  \end{subfigure}
  \begin{subfigure}{0.7\linewidth}
    \includegraphics[width=\linewidth]{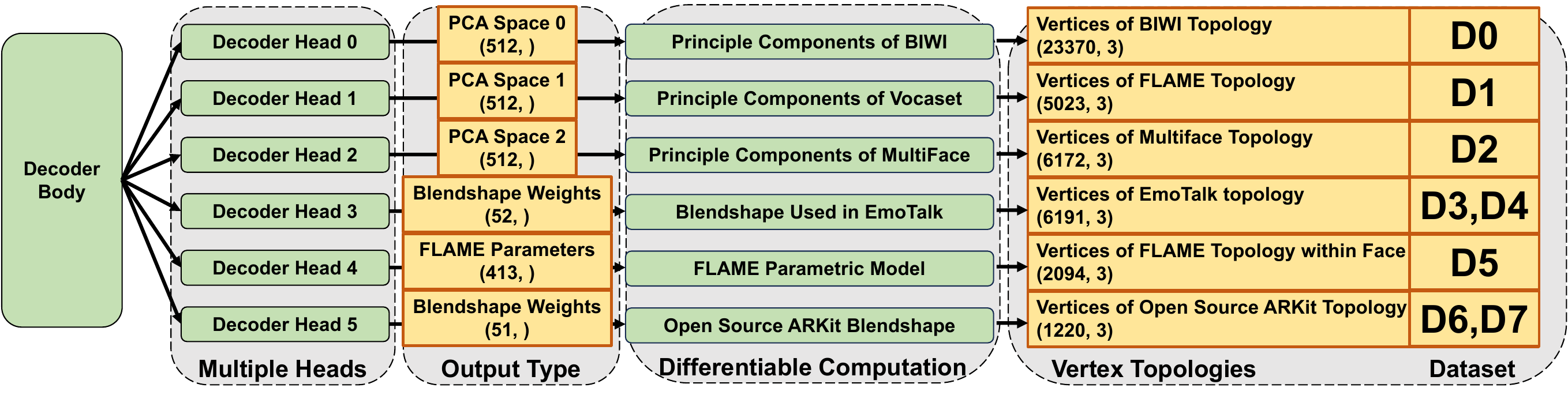}
    \caption{Zoomed-in View of UniTalker Decoder}
    \label{subfig:decoder_details}
  \end{subfigure}
  \caption{\textbf{Architecture Comparison.}
  (a) Vanilla multi-head audio-to-face model.
  (b) UniTalker adopts PCA to balance the annotation dimension across datasets, uses decoder warm-up to stabilize training, and develops a pivot identity embedding to mitigate dataset bias.
  (c) Zoomed-in view of UniTalker-[D0-D7] decoder. UniTalker-[D0-D7] has 6 decoder heads. 
  }
  \label{fig:model_architecture_comparison}
\end{figure}

A vanilla multi-head model is shown in \cref{subfig:unitalker_archi_vanilla}. The vanilla multi-head model doesn't lead to better results than the single-dataset-trained model. UniTalker model, as shown in \cref{subfig:unitalker_archi_unitalker} adopts PCA, DW strategies to improve the training stability as explained in Main Paper Sec 3.3, and PIE to mitigate dataset bias as explained in Main Paper Sec 3.2.
The detailed structure of UniTalker decoder trained on A2F-Bench, \ie, [D0-D7] is shown in \cref{subfig:decoder_details}. It has 6 decoder heads, corresponding to the 6 annotation types of the 8 datasets. 

\subsection{A2F-Bench Construction}
We have utilized five publicly available 3D audio-visual datasets, BIWI \cite{fanelli20103}, Vocaset \cite{cudeiro2019capture}, Multiface \cite{wuu2022multiface,richard2021meshtalk}, 3D-ETF-HDTF \cite{peng2023emotalk} and 3D-ETF-RAVDESS \cite{peng2023emotalk}. We created three additional datasets to enhance the model's proficiency in handling diverse languages and musical content. 

We cleaned and annotated the 2D faceforensics++ dataset \cite{rossler1901learning} and labeled the speaker's faces with FLAME \cite{li2017learning} parameters using 3D face reconstruction \cite{dad3dheads}. We collected a dataset consisting of recordings from eight native Chinese speakers and another dataset comprising recordings from eleven professional singers. A summary of the dataset information is provided in Main Paper Tab. 1. For simplicity, we refer to each dataset as D0, D1, ..., D7. 

We allocate the training, validation, and test sets in a ratio of 8:1:1. For BIWI (D0), we follow the processing pipeline in CodeTalker~\cite{xing2023codetalker}, which scales the vertices to  around [-0.5, 0.5]. For D1-D7, the coordinates are expressed in the measurement unit of meter. To balance the training of each decoder head, we duplicate the sequences in small datasets. The multiplication factors for D0 to D7 are 10, 5, 4, 1, 1, 1, 1, 1, respectively.

\noindent\textbf{BIWI (D0).}  The BIWI dataset consists of affective speech recordings paired with dense dynamic 3D face geometries. It includes a total of 40 sentences spoken by 14 subjects, comprising eight females and six males. Each sentence was recorded twice, once with emotion and once without. On average, each sentence has a duration of 4.67 seconds. The 3D face dynamics are captured at a frame rate of 25 fps, with each frame containing 23,370 vertices. To ensure fair comparison, we adopt the data splits used in previous studies \cite{fan2022faceformer,xing2023codetalker,peng2023selftalk}. The training set (BIWI-Train) consists of 190 sentences (2 sentences missing), while the validation set (BIWI-Val) comprises 24 sentences. There are two test sets: BIWI-Test-A, which includes 24 sentences spoken by six subjects seen during training, and BIWI-Test-B, which contains 32 sentences spoken by eight unseen subjects. 
In addition to the original 25 fps version, we interpolated the annotations to a frame rate of 30 fps, referred to as BIWI30, for experiments on the impact of PCA and DW strategies in paper Fig. 3. By using BIWI30 and Vocaset, we aim to eliminate the influence of mismatched frame rates on training stability. We use the original 25 fps version in all other experiments for fair comparison with prior works. 

\noindent\textbf{Vocaset (D1).} The Vocaset consists of 480 paired audio-visual sequences recorded from 12 subjects. Each sequence captures facial motion at a frame rate of 60 fps and has a duration of approximately 4 seconds. Unlike BIWI, the 3D face meshes in Vocaset are registered to the FLAME topology, resulting in meshes with 5,023 vertices. Previous studies such as FaceFormer \cite{fan2022faceformer}, CodeTalker \cite{xing2023codetalker} did not report quantitative results on Vocaset due to the absence of ground truth identity labels in the original test set. To ensure fair quantitative comparison, we have redivided the Vocaset and retrained their models using their official implementation as baseline. For each subject, we randomly split the 40 recorded sequences into training, validation, and test sets, comprising 32, 4, and 4 sequences, respectively. This new division allows for consistent evaluation and comparison of results across different models.

\noindent\textbf{Multiface (D2).} The Multiface dataset comprises high-quality recordings of the faces from 13 identities. The recordings were captured in a multi-view stage, capturing the subjects performing various facial expressions. On average, each subject has between 12,200 to 23,000 frames, with a capture rate of 30 frames per second. Following the approach described in MeshTalk \cite{richard2021meshtalk}, the meshes are transformed into the MeshTalk topology, resulting in meshes with 6,172 vertices.


\noindent\textbf{3DETF-HDTF (D3) and 3DETF-RAVDESS (D4).} The High-Definition Talking Face (HDTF) dataset is a collection of approximately 16 hours of videos sourced from YouTube.
The dataset includes recordings from over 300 subjects and includes around $10,000$ different sentences.
The RAVDESS dataset, short for the Ryerson Audio-Visual Database of Emotional Speech and Song, is a multi-modal emotion recognition dataset.
It consists of recordings from 24 actors with an equal split of 12 male and 12 female actors. The dataset comprises a total of $1,440$ video clips of short speeches, each accompanied by high-quality audio and video recordings. The actors were given specific instructions to express various emotions, including neutral, calm, happy, sad, angry, fearful, disgusted, and surprised. The speech content of RAVDESS only contains two utterances. In EmoTalk~\cite{peng2023emotalk}, a subset of the HDTF dataset consisting of five hours of videos was selected. This subset and  RAVDESS dataset were then labeled with 52 blendshape weights for each frame.


\noindent\textbf{Faceforensics++ (D5).} FaceForensics++ is a forensics dataset designed for evaluating face manipulation detection methods. It consists of 977 original videos sourced from YouTube. The original youtube videos contain trackable frontal face sequences without occlusions. We selected sequences from the original YouTube videos and split them into 1,714 sequences. These sequences were then labeled with FLAME parameters using DAD-Heads \cite{dad3dheads}.

\noindent\textbf{Our Speech(D6) and Song(D7) Datasets.} Our facial capture system utilizes ARKit with a depth camera on the iPhone 13 Pro to extract 51 blendshape weights, excluding TongueOut, at a frame rate of 60 fps. These blendshape targets are based on the widely-used Facial Action Coding System (FACS) and are suitable for industry novice users. Simultaneously, we record audios at a sample rate of 44,100 Hz.  Our speech dataset consists of 1.24 hours of Chinese speech recordings from 2 female and 6 male native Chinese speakers. Our song dataset comprises 5.11 hours of song recordings from 6 female and 5 male professional singers.

In the future, with improved estimation techniques~\cite{
zeng2022smoothnet,filntisis2022visual,chai2023hiface,wang2023zolly,zeng20233d,bolkart2023instant,sun2024aios,yin2024whac} or noise-robust learning schemes~\cite{veit2017learning,yang2020learn,chen2023understanding}, A2F-Bench could potentially incorporate more in-the-wild data for training and validation, scaling to include hundreds of hours of high-quality data.




\subsection{PCA Implementation}
We adopt PCA for each vertex-based annotation.
Specifically, for each vertex-based dataset, we assemble the annotations of all frames into a 2D matrix $\mX$ with shape  ${(F, 3V)}$, where $F$ represents the number of frames in all training sequences and $V$ denotes the number of vertices. 
%
$F \times 3V$ is usually too large for the solver to compute the principle components of $\mX$, thus leading to out-of-GPU-memory issue. 
Therefor, we employ Incremental Principal Components Analysis (I-PCA) \cite{ross2008incremental} with a batch size of $1024$ to incrementally approximate PCA components $\mW$. We retain the first $L=512$ principal components $\mW_L$ for each vertex-based annotation.
Before I-PCA, the 2D matrix $\mX$ is shuffled along the frame axis so that the distribution of vertices in each batch is approximately independent and identical.

\subsection{Training Loss}

To achieve balanced weight during the training of each annotation type, we apply vertex position scaling to ensure the vertices of each annotation are in a comparable range.
For BIWI dataset, we employ a fixed scaling factor of $0.2$, while for the Multiface dataset, we use a scaling factor of $0.001$. When dealing with parameter-based annotations, we perform scaling on the blendshape bases and the skeleton model. This approach ensures that the vertices from each head carry similar meaning in terms of measurement unit, which is meter after scaling. The reported LVE (Lip Vertex Error) for BIWI dataset is in the original space for fair comparison with prior works~\cite{fan2022faceformer,xing2023codetalker,peng2023selftalk,stan2023facediffuser}.
For blendshape weights in D3, D4, D6 and D7, we compute vertices according to \cref{eq:blendshape_to_verts}, 
\begin{equation}
    S_{face} = \Bar{S} + \sum_{i=1}^{B} \alpha_{i}s_i \label{eq:blendshape_to_verts}
\end{equation}
where $S_{face}$ denotes the face vertices, $\Bar{S}$ denotes the mean shape or neutral shape vertices, $\alpha_{i}$ denotes the $i$th element of blend-shape weights, $s_{i}$ denotes the $i$th shape base and $B$ denotes the number of shape bases.
For FLAME parameters in D5, we compute vertices according to \cref{eq:skeleton_to_verts},
\begin{equation}
    S_{face} = LBS(\Bar{T}, J, \Vec{\theta}, \mathcal{W}) \label{eq:skeleton_to_verts}
\end{equation}
where $\Bar{T}$ denotes the rest pose face vertices, $J$ denotes the joints of the skeleton, $\Vec{\theta}$ denotes the pose vector, and $\mathcal{W}$ denotes the blendweights of LBS \cite{li2017learning}.


In summary, we can compute vertices from each decoder head output and the computation is differentiable (\cref{subfig:decoder_details}). Consequently, the model can be optimized using the vertices MSE loss. 



\end{document}